\pdfoutput=1

\documentclass[11pt]{article}

\usepackage{ACL2023}

\usepackage{times}
\usepackage{latexsym}

\usepackage[T1]{fontenc}

\usepackage[utf8]{inputenc}

\usepackage{microtype}

\usepackage{inconsolata}

\usepackage{graphicx}
\usepackage{subfig}
\usepackage{tabu}
\usepackage{booktabs} 
\usepackage{enumitem}
\usepackage{nicefrac}
\usepackage{wrapfig} 
\usepackage{subfig}
\usepackage{xcolor}

\usepackage{pifont} 

\usepackage{amsmath}
\usepackage{amssymb}
\usepackage{mathtools}
\usepackage{amsthm}
\usepackage{bbm}

\usepackage{algorithm}
\usepackage[noend]{algorithmic}
\usepackage{xcolor}

\newcommand{\theory}[1]{\ensuremath{\mathcal{#1}}}
\newcommand{\facts}[1]{\ensuremath{\mathcal{#1}}}
\newcommand{\rules}[1]{\ensuremath{\mathcal{#1}}}
\newcommand{\goal}[1]{\ensuremath{\mathcal{#1}}}
\newcommand{\algo}{\textsc{Lambada}}
\newcommand{\proved}{\textsc{Proved}}
\newcommand{\disproved}{\textsc{Disproved}}
\newcommand{\unk}{\textsc{Unknown}}
\newcommand{\module}[1]{\emph{#1}}

\title{LAMBADA: Backward Chaining for Automated Reasoning in Natural Language}

\author{Mehran Kazemi, Najoung Kim, Deepti Bhatia, Xin Xu, Deepak Ramachandran \\
  Google Research \\
  \texttt{\{mehrankazemi, njkim, bhatiad, xxujasime, ramachandrand\}@google.com}}

\begin{document}
\maketitle
\begin{abstract}
Remarkable progress has been made on automated reasoning with natural text, by
using Language Models (LMs) and methods such as Chain-of-Thought and Selection-Inference. These techniques search for proofs in the forward direction from axioms to the conclusion, which suffers from a combinatorial explosion of the search space, and thus high failure rates for problems requiring longer chains of reasoning. The classical automated reasoning literature has shown that reasoning in the backward direction (i.e. from the intended conclusion to supporting axioms) is significantly more efficient at proof-finding. 
Importing this intuition into the LM setting, we develop a \emph{Backward Chaining} algorithm, called \algo, that decomposes reasoning into four sub-modules. These sub-modules are simply implemented by few-shot prompted LM inference.
We show that \algo\ achieves sizable accuracy boosts over state-of-the-art forward reasoning methods on challenging logical reasoning datasets, particularly when deep and accurate proof chains are required.
\end{abstract}

\section{Introduction}
Automated reasoning, the ability to draw valid conclusions from explicitly provided knowledge, has been a fundamental goal for AI since its early days \cite{mccarthy1959programs,hewitt1969planner}. Furthermore, logical reasoning, especially reasoning with unstructured, natural text is an important building block for automated knowledge discovery and holds the key for future advances across various scientific domains. While in recent years tremendous progress has been made towards natural language understanding thanks to pretrained language models (LMs) \cite[\textit{i.a.,}]{brown2020language,chowdhery2022palm}, the performance of these models for logical reasoning still lags behind \cite{rae2021scaling,creswell2023selection,valmeekam2022large} compared to the advancements in other areas such as reading comprehension and question-answering.

\begin{figure}[t]
  \centering
  \includegraphics[width=\columnwidth]{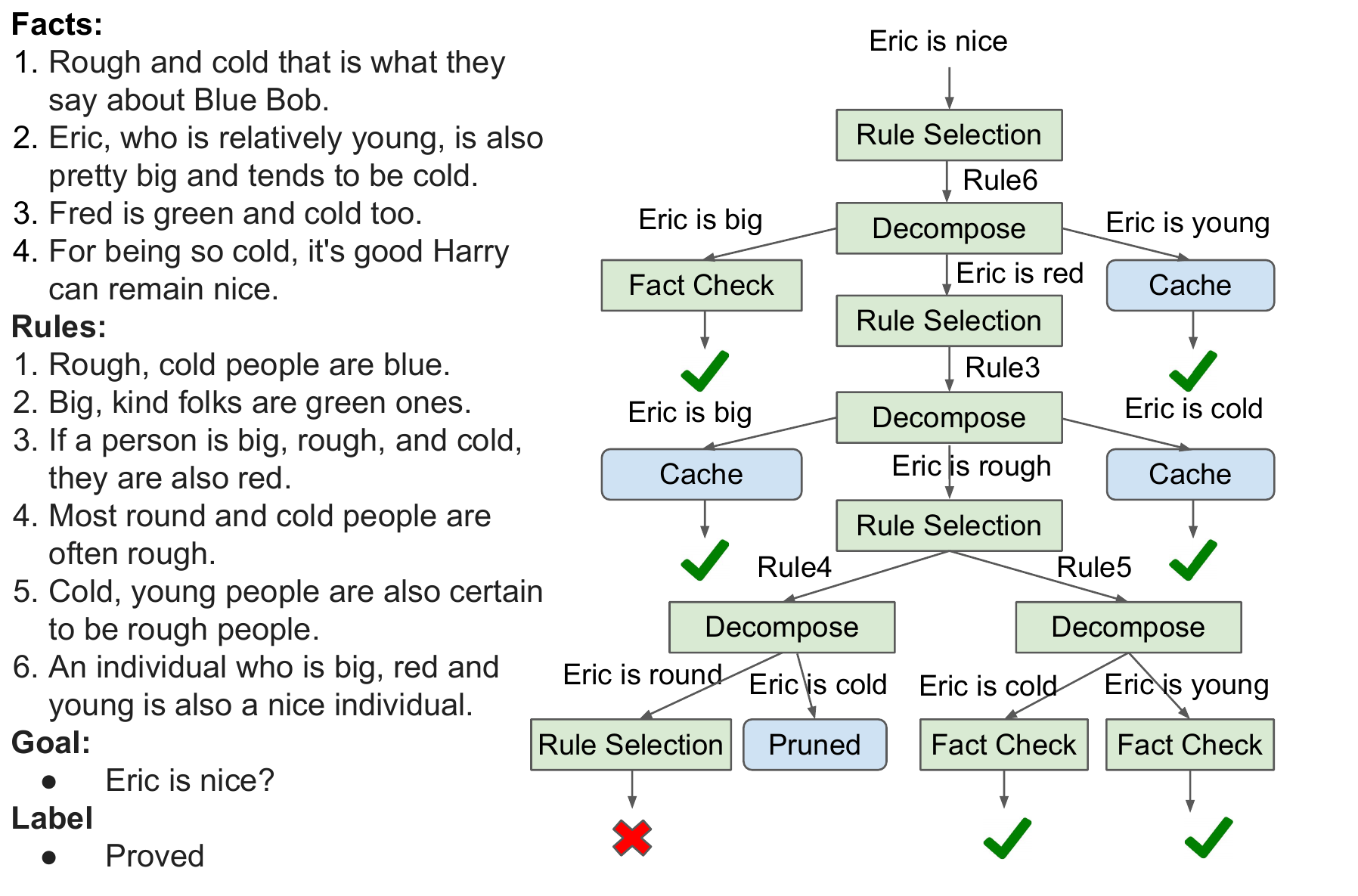}
  \caption{%
  \label{fig:success1} %
    The search trace of \algo\ on an example from the ParaRules subset of ProofWriter (the \module{Sign Agreement} and failed \module{Fact Check} modules are omitted for brevity). 
  }
\end{figure}

While many problems benefit from LM scaling, scaling has been observed to provide limited benefit for solving complex reasoning problems. For example, \citet{creswell2023selection} observed that for the Gopher family of LMs \citep{rae2021scaling}, the benefit of scaling for logic-based tasks is significantly worse than for other language tasks. 
Moreover, while finetuning initially seemed to enable logical reasoning in LMs \cite{clark2020transformers,tafjord2021proofwriter}, further exploration revealed that finetuned LMs mostly exploit spurious correlations 
(e.g., the correlation between the number of rules and the label) 
as opposed to learning to reason \cite{zhang2022paradox,schlegel2022can,liu2022transformers}.
Recently, prompting strategies such as Chain-of-Thought \cite{wei2022chain} and Scratchpad \citep{nye2022show} have contributed to improving performance of LMs on reasoning tasks, although they have been also shown to struggle with proof planning for more complex logical reasoning problems \cite{saparov2022language}.

One solution to the aforementioned problems is to integrate the strength and reliability of classical AI models in logical reasoning with LMs \cite{garcez2020neurosymbolic,marcus2020next}.
In the literature, there are two major approaches to logical reasoning \cite{poole2010artificial}: 
\begin{enumerate}[nosep,leftmargin=3.5mm]
    \item \emph{Forward Chaining (FC)} where one starts from the facts and rules (``theory''), and iterates between making new inferences and adding them to the theory until the goal statement can be proved or disproved,
    \item \emph{Backward Chaining (BC)} where one starts from the goal and uses the rules to recursively decompose it into sub-goals until the sub-goals can be proved or disproved based on the theory.
\end{enumerate}  
Previous approaches to reasoning with LMs mostly incorporate elements of FC into LMs \cite{tafjord2021proofwriter,creswell2023selection}. FC requires selecting a subset of facts and rules from the entire set, which might be difficult for an LM as it requires a combinatorial search over a large space. 
Moreover, deciding when to halt and declare failure to prove is challenging in FC, as also noted by \citet{creswell2023selection}, sometimes requiring specialized modules trained on intermediate labels \cite{creswell2022faithful}. Indeed, the classical automated reasoning literature is heavily weighted towards BC or goal-directed strategies for proof-finding. 

In this paper, we show experimentally that BC is better suited for text-based deductive logical reasoning, as it does not require a combinatorial search for subset selection and there are more natural halting criteria for it. We develop a hybrid \textbf{LA}nguage \textbf{M}odel augmented \textbf{BA}ckwar\textbf{D} ch\textbf{A}ining technique (\emph{\algo}), where BC drives the high-level proof planning, and the LM performs the textual understanding and individual reasoning steps.
We conduct experiments with challenging datasets for LM reasoning containing examples expressed in naturalistic text. The datasets contain proof chains of up to $5$ hops in depth, and examples where the goal can neither be proved nor disproved from the provided theory. 
We show that \algo\ achieves substantially higher deductive accuracy, and is considerably more likely to generate valid reasoning chains compared to other techniques which find correct conclusions with spurious proof traces, while also being more query efficient than other LM-based modular reasoning approaches. Our results strongly indicate that future work on reasoning with LMs should incorporate backward chaining or goal-directed planning strategies.

\section{Related Work}
The deep learning based models that have been developed to solve text-based (logical) reasoning tasks can be categorized as follows (see \citealt{huang2022towards} for a recent survey of the literature).

\textbf{Pretraining on Relevant Tasks:} Pretraining an LM on corpora relevant to the target reasoning task can lead to improvements \cite{hendrycks2021measuring,shen2021generate}. Pretraining is, however, costly especially for larger LMs. 

\textbf{Implicit Reasoning:} These approaches finetune LMs to produce the label directly given the input \cite{clark2020transformers,betz2020critical,saeed2021rulebert,han2022folio}; reasoning is expected to happen implicitly in the parameters of the LM.
It has been shown that finetuning LMs on logical reasoning tasks makes them learn spurious correlations \cite{zhang2022paradox,schlegel2022can}, and is not robust to multi-hop reasoning \cite{kassner2020pretrained}. Besides, finetuning large LMs is costly especially when the dataset is large, and may introduce distributional shocks to the model \cite{kazemi2023understanding}. In this paper, we focus on models that only take in-context examples as supervision.

\textbf{Explicit Reasoning:} Generating the intermediate reasoning steps such as the chain of reasoning \cite{wei2022chain,nye2022show,dalvi2021explaining,zelikman2022star,zhang2022improved} has shown substantial improvement for many reasoning tasks \cite{suzgun2022challenging}. Such chains have been explored both in the forward and the backward directions, e.g., using multiple constrained 
LMs for logical reasoning \citep{zhang2022improved}. \citet{gontier2020measuring} investigated how transformer models perform when trained to perform forward or backward chaining, and drew conclusions about their internal reasoning strategies. We compare against a popular recent prompting strategy, namely Chain-of-Thought (CoT) \cite{wei2022chain}, from this category. 

\textbf{Verifiers:} To improve CoT, some works train a verifier using chain-level labels. The verifier takes a reasoning chain produced by the model as input and judges the quality of the chain \cite{cobbe2021training,shen2021generate,jhamtani2020learning,zelikman2022star}. Using this verifier, one can then generate multiple reasoning chains (e.g., by running the algorithm multiple times with different decoding temperatures) and use the best chain according to the verifier. Since \algo\ also generates proofs, verifiers are also applicable to our algorithm. 
In this paper, we assume not having access to chain-level labels, and leave experiments with verifiers as future work.

\textbf{Length generalization:} A number of approaches specifically look into whether LMs can generalize from examples requiring shorter reasoning chains (shown to them either as demonstration or as finetuning data) to examples requiring longer chains \cite{anil2022exploring,tafjord2021proofwriter}. With our model, length generalization comes for free because the model learns the building blocks of solving the problem that are applied as many times as needed to solve the problem. 

\textbf{Modular Reasoning:} These approaches break the problem into smaller modules and use separate LMs to solve each module \cite{zhou2022teaching,khot2022decomposed,sprague2022natural,zhou2022least,dua2022successive,wang2022iteratively,schlag2023large}. 
LM-based approaches to logical reasoning typically makes use of a single LM module; for example, in \citet{tafjord2021proofwriter}, a single LM module iteratively and exhaustively infers \emph{all} conclusions based on the facts and rules, and then the goal statement is compared against the final set of conclusions to confirm if it can be proved from the theory. Since exhaustively deriving all conclusions is computationally expensive, \citet{creswell2023selection} consider a more scalable approach where the conclusions that are derived are informed by the goal; they iteratively apply two LLM modules one selecting a subset of the facts and rules informed by the goal and the other making new inferences based on the selected facts and rules and adding it back to the theory. In this paper, we compare against the second approach.

\textbf{Natural Language Inference (NLI):} Logical reasoning can also be understood as identifying whether a logical entailment relation holds between two propositions (premise and hypothesis; the premise is the theory and the hypothesis is the statement to be proved). In this sense, NLI models are also relevant, although inferences under NLI typically adopt a more relaxed notion of entailment rather than purely logical \cite{dagan2013recognizing,bowman2015large,N18-1101}.

\section{\algo: Language Model Augmented Backward Chaining} \label{sec:method}
We focus on performing automated reasoning over \emph{facts}, i.e., natural language  assertions such as \texttt{``Nice people are red''}, that are coherent but not necessarily grounded in reality. A \emph{rule} is a natural language statement that is either of the form, or can be rewritten in the form, \texttt{``If P then Q''}; e.g., \texttt{``Rough, cold people are blue''} can be rewritten as  \texttt{``If a person is rough and cold, then they are blue''}. \texttt{P} is called the \emph{antecedent} and \texttt{Q} is called the \emph{consequent} of the rule. A \emph{theory} \theory{C} consists of facts $\facts{F}=\{f_1, f_2, \dots, f_n\}$ and rules $\rules{R}=\{r_1, r_2, \dots, r_m\}$. We let \goal{G} represent a \emph{goal} that we would like to prove or disprove based on the theory. 
An example theory with fictional characters and rules is demonstrated in Figure~\ref{fig:success1}. 
Based on the theory, one should prove or disprove the goal \texttt{``Eric is nice''}.

\subsection{Backward Chaining}
Backward chaining (BC) is a strategy for reasoning that starts from the goal and recursively breaks the goal into sub-goals based on the rules that can be applied to it, until the sub-goals can be proved or disproved based on the facts or no more rules can be applied to break down the sub-goal further. 

Figure~\ref{fig:success1} shows an example of BC applied to a theory to prove a goal. Initially, BC verifies if the goal can be proved or disproved based on the facts (this step is omitted from the figure). Since none of the facts directly prove or disprove the goal, BC next selects a rule that can be applied to break down the goal into sub-goals. Whether or not a rule applies to a goal is determined by an operation called \emph{unification} in logic; Rule6 has the same consequent as the goal so the operation can be applied, but the other rules have different consequents and it cannot be applied. Using Rule6, 
the goal can be broken down into three sub-goals that should be proved for the goal to be proved. BC then makes recursive calls to prove each sub-goal. The algorithm continues until either a halting criterion is reached (e.g., reaching a certain depth in search), or a sub-goal can no longer be broken down (e.g., the left sub-tree under \texttt{``Eric is rough''}), or all sub-goals are proved (e.g., the right sub-tree under \texttt{``Eric is rough''}).

The outcome of BC for a goal is either \proved, \disproved, or \unk; e.g., its output for the goal in Figure~\ref{fig:success1} is \proved, for \texttt{``Fred is not green?''} is \disproved\ (because it contradicts Fact3), and for \texttt{``Fred is round?''} is \unk\ (because the theory does not entail or contradict it).

\subsection{LM Modules in \algo}
To enable applying BC for text-based reasoning, we introduce four LM-based modules: \emph{Fact Check}, \emph{Rule Selection}, \emph{Goal Decomposition}, and \emph{Sign Agreement}, each implemented by showing relevant in-context demonstrations to a pretrained LM (see Appendix~\ref{sec:prompts} for details). We describe these modules and then proceed to the full algorithm.

\subsubsection{Fact Check} 
Given a set of facts \facts{F} from the theory and a goal \goal{G}, the \module{Fact Check} module verifies if there exists a fact $f\in\facts{F}$ such that $f$ entails \goal{G} (in which case the goal is proved) or $f$ entails the negation of \goal{G} (in which case the goal is disproved). If no such fact can be found, then the truth of \goal{G} remains unknown. 

We implement \module{Fact Check} with two sub-modules: the first sub-module selects a fact from the set of facts that is most relevant to the goal, and the second sub-module verifies if the goal can be proved or disproved based on that fact.\footnote{Note that we select only one fact because the goals and sub-goals in the datasets we work with can be proved/disproved using single facts; The two modules can be adapted to selected multiple facts if this is not the case.}
Since the first sub-module may fail to identify the best fact on the first try, if the truth of the goal remained unknown after one try, the selected fact can be removed and the sub-modules can be called again. This process can be repeated multiple times. In our experiments, we call the two sub-modules twice. 

\subsubsection{Rule Selection}
Given a set of rules \rules{R} from the theory and a goal \goal{G}, the \module{Rule Selection} module identifies the rules $r\in\rules{R}$ such that the consequent of $r$ unifies with \goal{G}. These rules are then used for decomposing the goal into sub-goals. If no such rule can be identified, then the truth of \goal{G} remains unknown.

As we did for \module{Fact Check}, we implement \module{Rule Selection} with two sub-modules: the first sub-module identifies the consequent of each rule (independent of the goal), and the second sub-module takes the rule consequents and the goal as input and identifies which one unifies with the goal. 
Note that due to the recursive nature of BC, the \module{Rule Selection} module may be invoked multiple times during the proof of a goal. Since identifying the consequent of each rule is independent of the goal, this sub-module only needs to be called once.

\subsubsection{Goal Decomposition}
Given a rule $r$ and a goal \goal{G} such that the consequent of $r$ unifies with \goal{G}, the \module{Goal Decomposition} module identifies the sub-goals that need to be proved in order for $\goal{G}$ to be proved or disproved. The sub-goals are identified based on the antecedent of $r$.

\subsubsection{Sign Agreement}
In the case where we succeed in proving the antecedent of $r$, whether the goal is proved or disproved depends on whether the sign of the goal agrees or disagrees with the sign of the consequent of $r$. For instance, in Figure~\ref{fig:success1}, for the goal \texttt{``Eric is nice.''}, since the sign of the goal agrees with the sign of the consequent of Rule6 and the antecedent of the rule is proved, we conclude that the goal is proved. However, if Rule6 was \texttt{``[...] is not going to be a nice individual.''}, then the sign of the goal would disagree with the sign of the consequent and so we would conclude that the goal is disproved. This motivates the fourth module, \module{Sign Agreement}, described below.

Given a rule $r$ and a goal \goal{G}, the \module{Sign Agreement} module verifies if the sign of the consequent of $r$ agrees or disagrees with the sign of the goal or not.

\begin{algorithm}[t]
\caption{\algo}
\label{algo:back-chain}
\textbf{Input:} Theory $\theory{C}=(\facts{F}, \rules{R})$, Goal $\goal{G}$, Max-Depth D
\begin{algorithmic}[1]
\STATE factCheckResult = \textit{\textcolor{blue}{FactCheck}}(\goal{G}, \facts{F})
\IF{factCheckResult $\neq$ \unk}
    \STATE \textbf{return} factCheckResult
\ENDIF
\IF{D == 0}
    \STATE \textbf{return} \unk
\ENDIF
\STATE $\rules{R}_s$ = \textit{\textcolor{blue}{RuleSelection}}(\goal{G}, \rules{R})
\FOR{$r \in~$Rerank$(\rules{R}_s$)}
    \STATE \textbf{\goal{\mathbf{G}}} = \textit{\textcolor{blue}{GoalDecomposition}}($r, \goal{G}$)
    \IF{ProveSubgoals(\theory{C}, \goal{\mathbf{G}}, D)}
        \IF{\textit{\textcolor{blue}{SignAgreement}}(r, \goal{G})}
            \STATE \textbf{return} \proved
        \ELSE
            \STATE \textbf{return} \disproved
        \ENDIF
    \ENDIF
\ENDFOR    
\STATE \textbf{return} \unk

\end{algorithmic}
\end{algorithm}

\begin{algorithm}[t]
\caption{ProveSubgoals}
\label{algo:prove-sub-goals}
\textbf{Input:} Theory $\theory{C}=(\facts{F}, \rules{R})$, Sub-Goals $\goal{\mathbf{G}}$, Max-Depth D
\begin{algorithmic}[1]
\FOR{\goal{G} in \goal{\mathbf{G}}}
    \STATE result = \algo(\theory{C}, \goal{G}, D-1)
    \IF{result $\neq$ \proved}
        \STATE \textbf{return} False \emph{\# Assuming conjunction}
    \ENDIF
\ENDFOR

\STATE \textbf{return} True
\end{algorithmic}
\end{algorithm}

\subsection{The \algo\ Algorithm}
Algorithm~\ref{algo:back-chain} provides a high-level description of how the four LM modules described earlier can be integrated with BC to enable text-based logical reasoning (the function calls corresponding to LM modules are color-coded). 

\algo\ can be understood as a depth-first search algorithm over the facts and the rules. It takes as input a theory $\theory{C}=(\facts{F}, \rules{R})$, a goal \goal{G}, and a depth $D$ that defines a halting criterion for the algorithm based on the maximum allowed depth for the search. The search depth is a natural halting criterion corresponding to the maximum number of reasoning hops required for answering questions.

Initially, the algorithm uses the \module{Fact Check} module to check if \goal{G} can be proved or disproved using the facts. If this is the case, then the algorithm stops and returns the result (\proved\ or \disproved). 

If \goal{G} cannot be proved or disproved, then the algorithm checks the depth $D$: if $D=0$, then the algorithm stops and returns \unk\ indicating that \goal{G} could not be proved or disproved. Otherwise, the algorithm proceeds with applying rules.

The \module{Rule Selection} module is used to identify the rules $\rules{R}_s$ from $\rules{R}$ whose consequent unifies with \goal{G}. Once the set $\rules{R}_s$ is identified, if \algo\ can start with the rules that have a higher chance of succeeding at (dis)proving the goal, it can save computations and be less error-prone. Therefore, we include a \emph{Rerank} function in \algo. Based on the intuition that shorter rules are likely to have fewer sub-goals (hence a higher chance of success), we start the search from shorter rules and proceed to longer rules if the shorter ones fail. 
We leave more sophisticated ranking strategies as future work.

For each selected rule, the algorithm uses the \module{Goal Decomposition} module to decompose \goal{G} into a set of sub-goals $\mathbf{G}$ that need to be proved and checks whether those sub-goals can be proved by making recursive calls to the algorithm (with reduced depth). If the sub-goals can be proved, then the algorithm uses the \module{Sign Agreement} module to check whether the sign of the rule consequent agrees or disagrees with the sign of \goal{G}. If it does, then the algorithm returns \proved\ and otherwise \disproved. If there is no rule for which the sub-goals can be proved, then \unk\ is returned.

During a proof, \algo\ may be called multiple times with the same theory and goal; in Appendix~\ref{sec:cache} we explain how cycles and redundant computations can be avoided using a cache.

\begin{figure*}[t]
  \centering
  \subfloat[ProofWriter-PUD]{%
  \includegraphics[width=0.8\columnwidth]{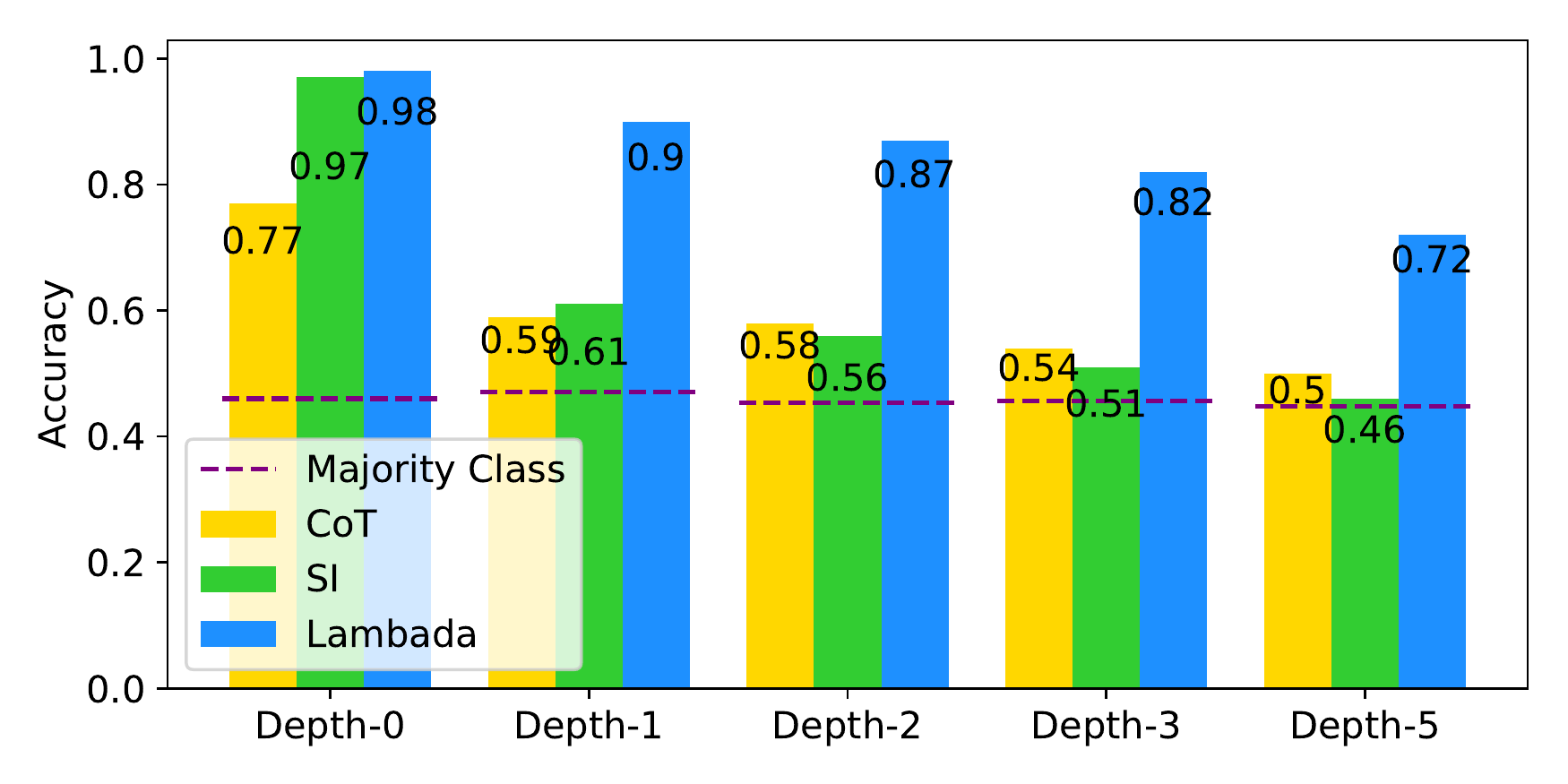}}
~~~~\hspace*{1cm}
  \subfloat[ProofWriter-PD]{%
  \includegraphics[width=0.8\columnwidth]{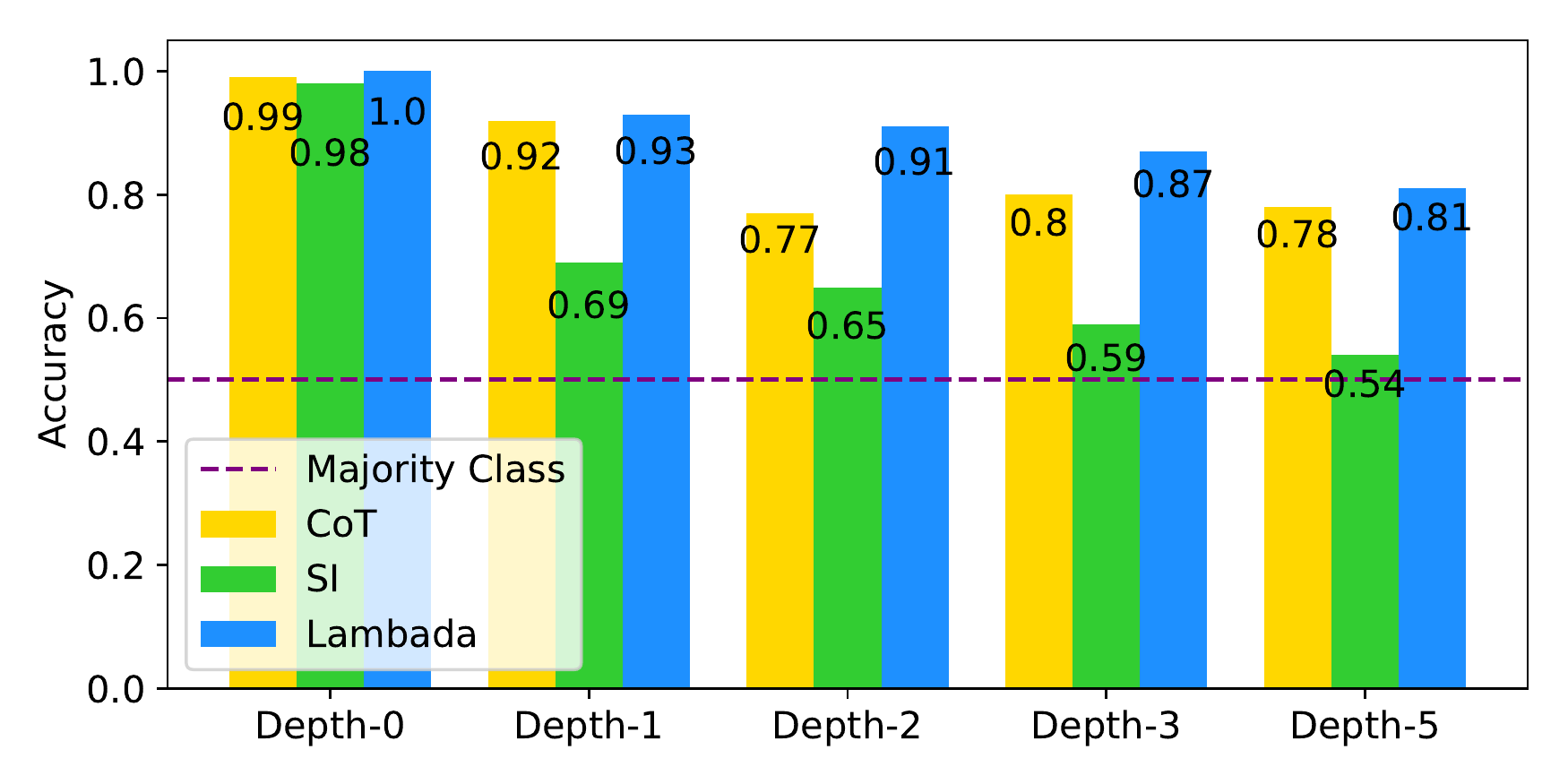}} %
  \\
  ~~~~\hspace*{-1.0cm}
  \subfloat[PrOntoQA]{%
  \includegraphics[width=0.8\columnwidth]{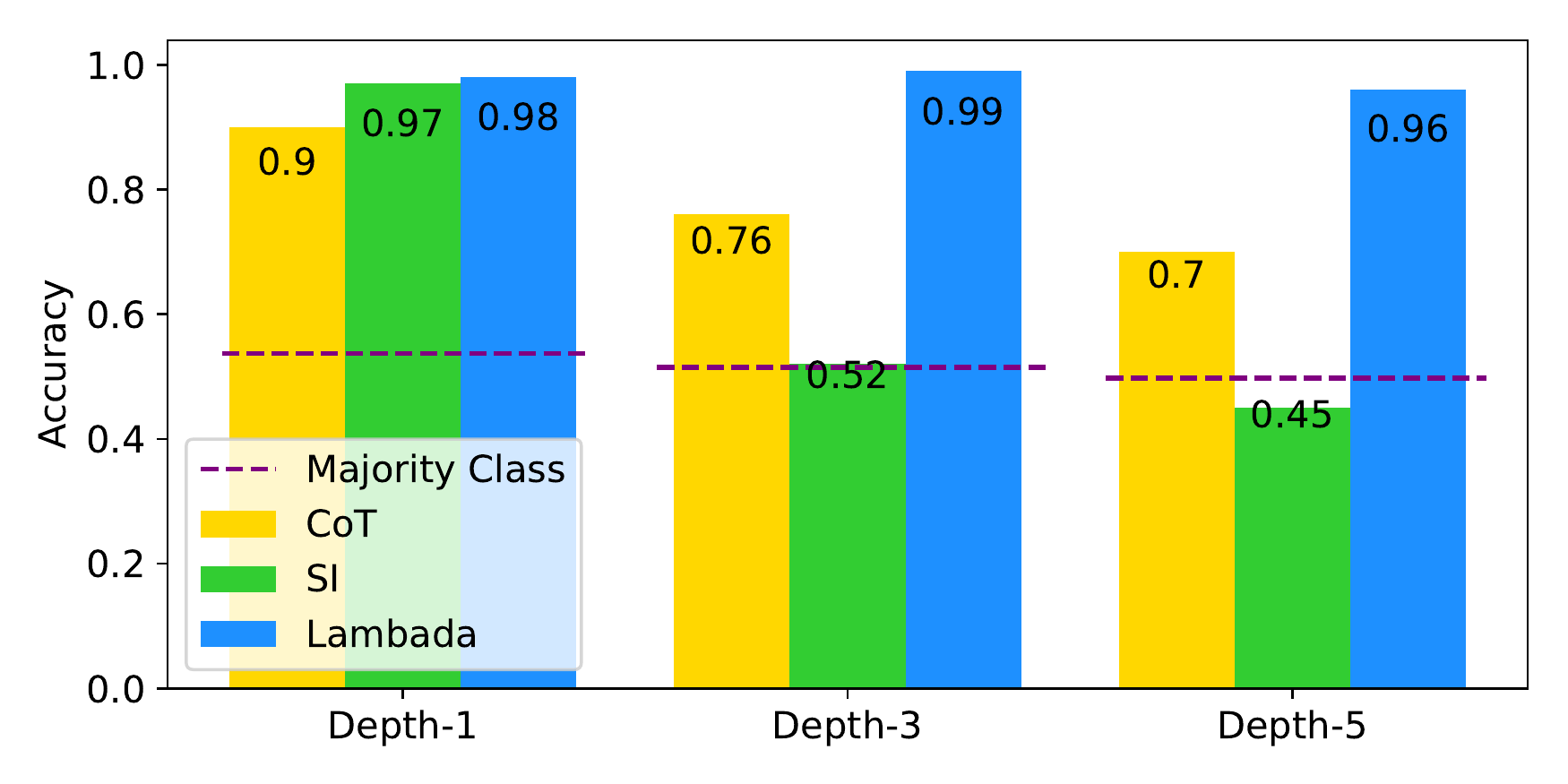}} %
  ~~~~\hspace*{1cm}
  \subfloat[ParaRules]{%
  \includegraphics[width=0.3\columnwidth]{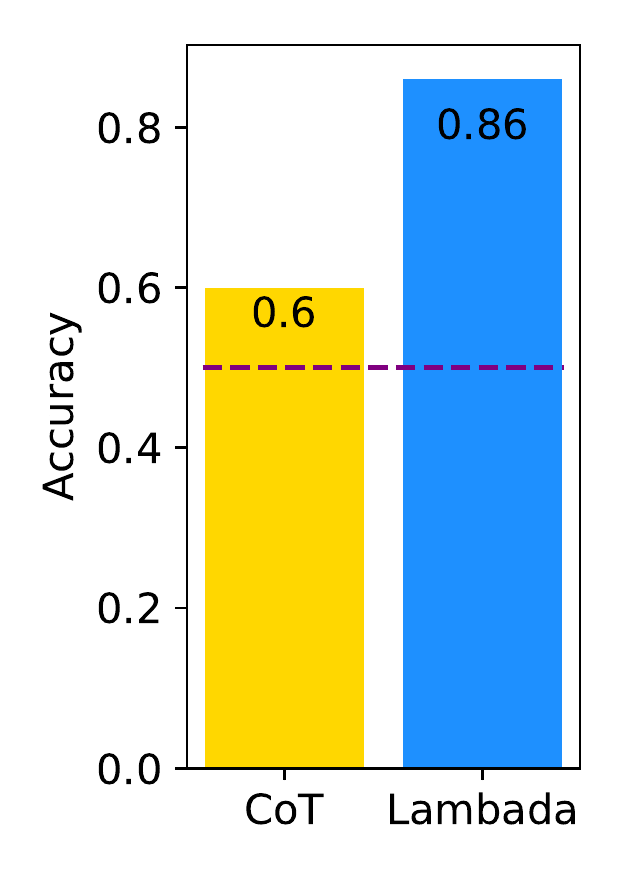}} %
  ~~~~\hspace*{1cm}
  \subfloat[ProofWr.~(d5)]{%
  \includegraphics[width=0.3\columnwidth]{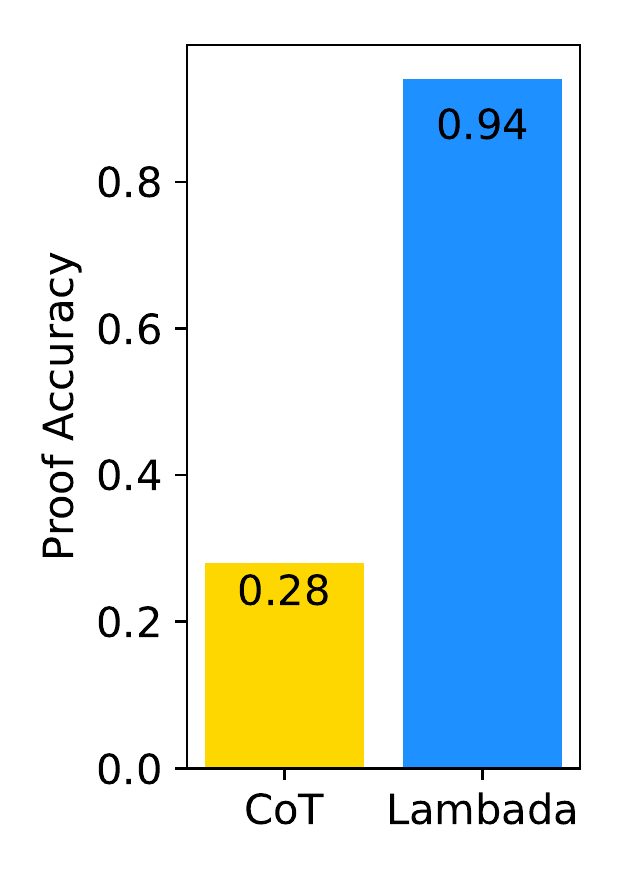}} %

  \caption{%
  \label{fig:main-results} %
  Prediction accuracy results on (a) ProofWriter-PUD (b) ProofWriter-PD, (c) PrOntoQA, and (d) ParaRules datasets. (e) The proof accuracy of CoT and \algo\ on ProofWriter (Depth-5) for a set of randomly sampled examples for which the models correctly predicted if the goal can be proved or disproved.}
\end{figure*}

\section{Experimental Setup}
We describe our baselines and datasets here, and provide further implementation details in Appendix~\ref{sec:impl_details}. Unless stated otherwise, all experiments are based on the PaLM 540B model \cite{chowdhery2022palm}. 

\subsection{Baselines} 
We compare against the following two baselines.

\textbf{Chain-of-Thought (CoT)} \cite{wei2022chain} is a popular neural approach based on demonstrating chains of inference to the LM within the in-context prompt. In addition to the few-shot demonstrations in \texttt{<INPUT>/<LABEL>} format in typical in-context learning settings, in CoT, an intermediate explanation for the label is also provided (\texttt{<INPUT>/<EXPLANATION>/<LABEL>}). In our work, the explanation corresponds to the proof.

\textbf{Selection-Inference (SI)} \cite{creswell2023selection} is a strong modular reasoning approach based on forward chaining. SI contains two modules: (1) \emph{selection}, which, guided by the goal, selects a subset of the facts and rules from which new conclusions can be derived toward proving the goal, and (2) \emph{inference}, which takes the selected facts and rules and derives a new conclusion. The two modules are called iteratively, each time producing a single conclusion that is added back to the theory before the next iteration. The iterations continue until a halting criterion is met (a fixed number of steps in \citealt{creswell2023selection}).

\subsection{Datasets}
We experiment with challenging deductive logical reasoning datasets outlined below.

\textbf{ProofWriter} \cite{tafjord2021proofwriter} is a commonly used synthetic dataset for testing logical reasoning when facts and rules are expressed in naturalistic text. It contains two subsets: an open-world assumption (OWA) subset and a closed-world assumption (CWA) subset. In this paper, we use the OWA subset. Each example is a (\textit{theory, goal}) pair and the label is one of $\{$\proved, \disproved, \unk$\}$ where \unk\ indicates that the goal can neither be proved nor disproved. The dataset has five parts, each part requiring $0$, $\leq 1$, $\leq 2$, $\leq 3$ and $\leq 5$ hops of reasoning, respectively. We report two sets of results on this dataset: (1) with examples labeled \unk\ removed (for compatibility with previous work), and (2) with all three labels. Note that intermediate proof chains from ProofWriter are not used by our models in making predictions. For both cases, due to the cost of inference, we used the first $1000$ examples in the test set. Hereafter, we refer to these two subsets as \emph{ProofWriter-PD} and \emph{ProofWriter-PUD}. 

\textbf{PrOntoQA} \cite{saparov2022language} is a synthetic dataset created to analyze the capacity of LM-based approaches for logical reasoning. Compared to ProofWriter, PrOntoQA has lower natural language diversity and less l fact/rule variations (e.g., no conjunctions). However, the search traces typically contain multiple paths with only one of them leading to the proof, thus enabling testing the proof planning of different models. This dataset has multiple versions; we use the \emph{fictional characters} version, which is one of the hardest versions according to \citet{saparov2022language}. Similarly to ProofWriter, each version of PrOntoQA is divided into different parts depending on the depth of reasoning chains required ($1$, $3$, and $5$ hops).

\textbf{ParaRules} \cite{tafjord2021proofwriter} is a version of ProofWriter where the synthetically generated sentences in the theory are rewritten by crowdworkers to increase diversity and naturalness of the text. This lets us move beyond evaluating reasoning with templatic expressions, which is a key limitation of the other datasets. Each fact in ParaRules may be a combination of several sub-facts (see Fig.~\ref{fig:success1} for an example). The examples require proof depths of up to $5$ and the label can be \proved, \disproved, or \unk. We found some minor quality issues in ParaRules; we manually verified and fixed the first $500$ examples of the test set (see Appendix~\ref{sec:natlang-preproc}) and used this set for evaluation.

\section{Results}
We now describe the results and compare \algo\ and the baselines in detail.

\subsection{Label Prediction Accuracy}
The results are reported in Figure~\ref{fig:main-results}, (a)--(d).\footnote{Due to the low performance of SI on ProofWriter and PrOntoQA and its high number of LM calls (see Figure~\ref{fig:call_count}), we only compared \algo\ against CoT for ParaRules.} \algo\ significantly outperforms the baselines, especially on ProofWriter-PUD which contains \unk\ labels ($44\%$ relative improvement compared to CoT and $56\%$ compared to SI on Depth-5), the higher depths of PrOntoQA ($37\%$ relative improvement compared to CoT and $113\%$ compared to SI on Depth-5), and the ParaRules dataset ($43\%$ relative improvement compared to CoT). These results overall show the merit of \algo\ for logical reasoning. We highlight that the reasoning capacity of \algo\ robustly generalizes to more naturalistic expressions, as demonstrated by the high accuracy on ParaRules, which is exactly the desired outcome of combining the strengths of an LM and a symbolic reasoning algorithm.

The results in Figure~\ref{fig:main-results}(a) reveal a shortcoming of the CoT approach in dealing with \unk\ labels. That is, unlike the examples for which the label is \proved\ or \disproved, there is no natural chain of thought for the examples whose labels are \unk.
Nevertheless, the performance of CoT is competitive for the ProofWriter-PD dataset, and the accuracy does not diminish substantially with increasing depth. We investigate the reason for this behaviour of CoT in the next section.

\begin{figure}[t]
  \centering
  \includegraphics[width=0.8\columnwidth]{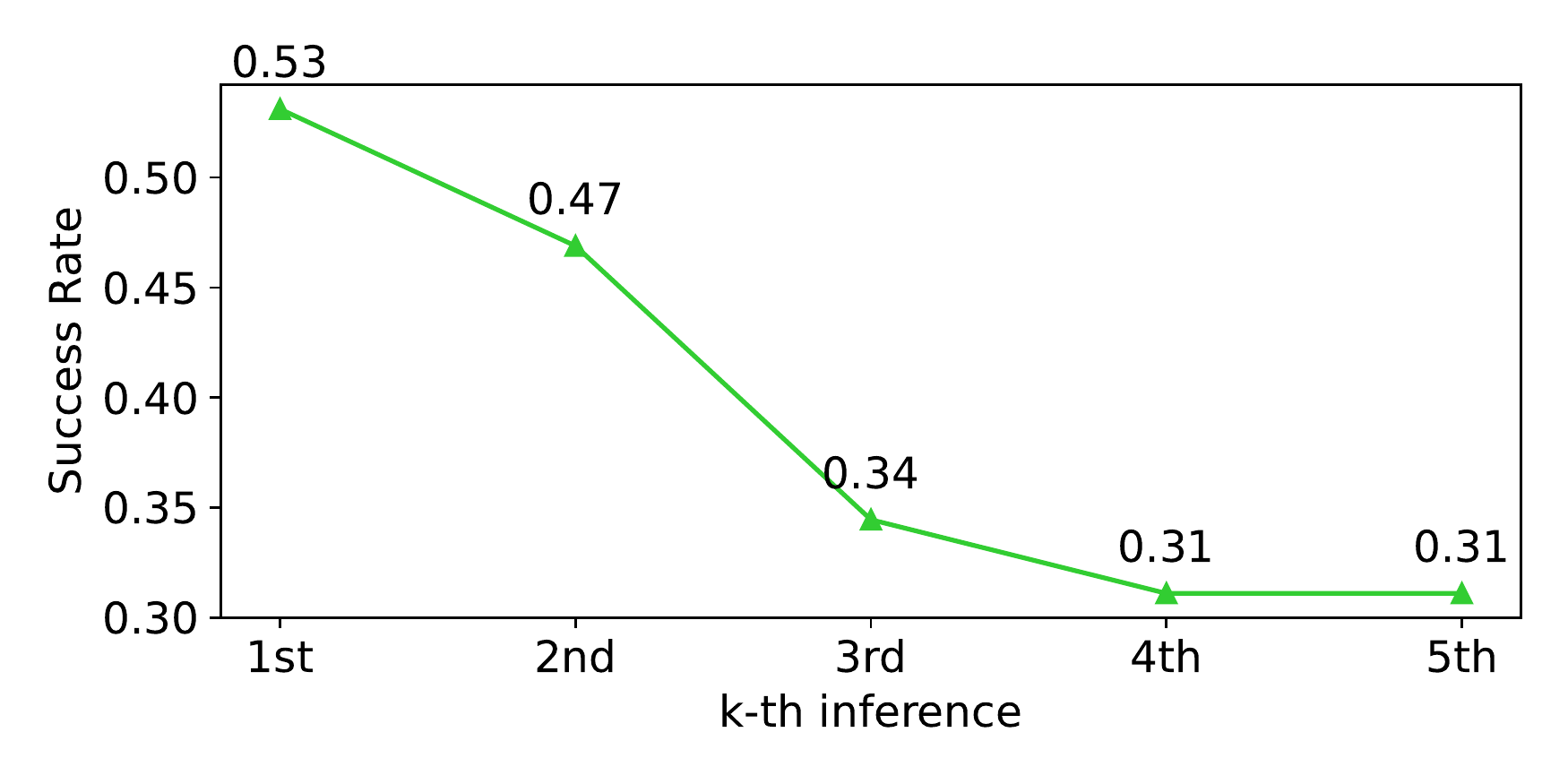}
  \caption{%
  \label{fig:si-becomes-harder} %
    The success rate of the $k$-th inference of SI on PrOntoQA (Depth-5) for different values of $k$. As $k$ increases, the size of the input theory becomes larger and the success rate decreases.
  }
\end{figure}

\subsection{Proof Accuracy}
To understand the reason behind the high accuracy of CoT on higher depths of ProofWriter-PD, we randomly selected $50$ examples from Depth-5 of the dataset where CoT predicted the label correctly, and manually verified if the proof chain is correct or not. For comparison, we also manually verified the proofs generated by $\algo$ following a similar procedure. The results are reported in Figure~\ref{fig:main-results}(e). 

While \algo\ mostly produces correct chains, CoT produces correct chains only for $28\%$ of the examples. We find that hallucination is the main source of error ($48\%$ of the examples; see Appendix~\ref{sec:cot-proof-errors} for other prominent failure modes). The hallucinated facts and rules mostly resulted in shortcuts to the correct answer. This hints at the possibility of spurious correlations in ProofWriter-PD that can be exploited by CoT (see Appendix~\ref{sec:cot-proof-errors}, Figure~\ref{fig:cot-failures} for examples). This result is consistent with previous work showing that when LMs are asked to solve logical reasoning end-to-end, they rely on spurious correlations \cite{zhang2022paradox}. Note that for modular approaches like SI and \algo, the intermediate modules are impervious to the spurious correlations between the input and the label and do not suffer from this issue.

\begin{figure}[b]
  \centering
  \includegraphics[width=0.7\columnwidth]{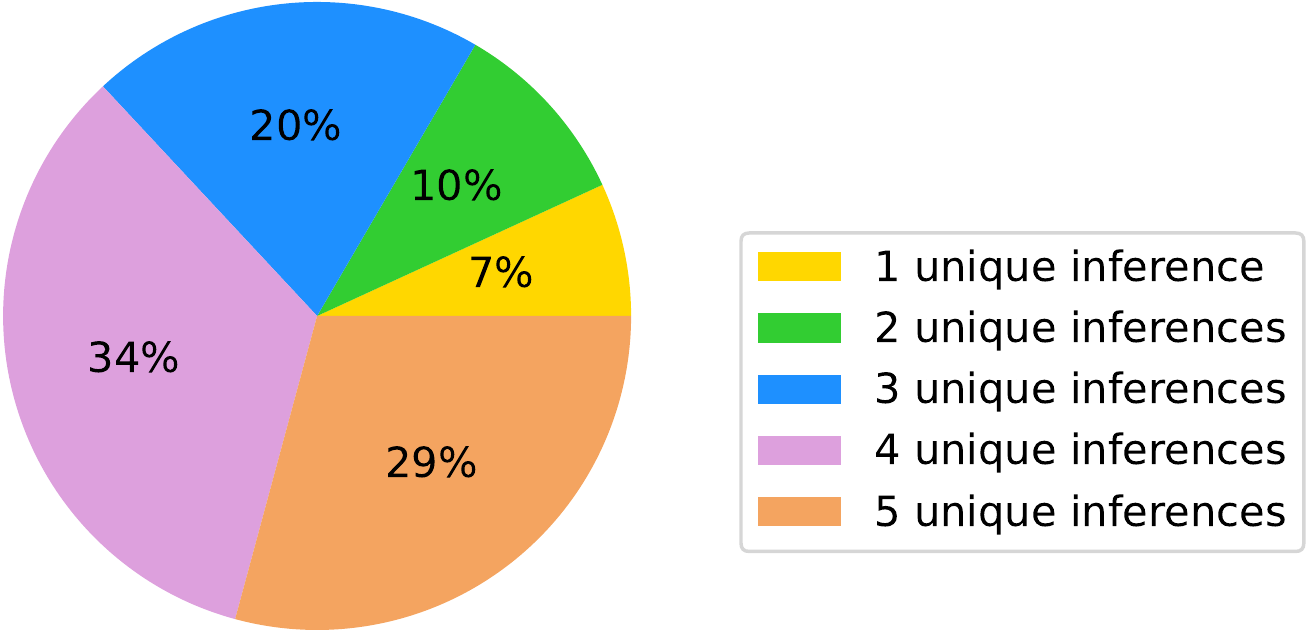}
  \caption{%
  \label{fig:si-dup} %
    Number of unique inferences generated by SI for Depth-5 of ProofWriter-PUD when selection and inference modules are called five times.
  }
\end{figure}

\subsection{Forward vs. Backward Chaining}
As previously explained, SI is based on forward chaining and its selection module requires a combinatorial search to find the right subset of facts and rules (see Appendix~\ref{sec:combinatorial-search}), and the search space becomes progressively larger in each iteration of the algorithm as new inferences are added to the theory. 
To verify whether the increase in the search space makes forward chaining progressively harder, we measured the success rate of the $k$-th inference of SI for different values of $k$ on Depth-5 of PrOntoQA (see Appendix~\ref{sec:si-becomes-harder} for details). From the results in Figure~\ref{fig:si-becomes-harder}, we can see that the success rate indeed decreases in the later inferences of the model, where the size of the input theory is larger and therefore a larger space needs to be searched to find the right combination of facts and rules.
Note that none of the components in \algo\ require selecting a \emph{subset}, hence no combinatorial search is required (see Appendix~\ref{sec:combinatorial-search} for more details).

SI also suffers from inferring redundant facts. Figure~\ref{fig:si-dup} reports the number of unique inferences from SI for the examples in ProofWriter-PD (Depth-5) where SI incorrectly predicted \unk\ (i.e., examples where a proof exists but SI failed to find it). The result shows that SI inferences contained no redundant facts only $29\%$ of the time; in $7\%$ of the cases, all $5$ inferred facts were identical, and in another $10\%$, only two unique inferences were made. This shows that SI, and maybe more generally forward-chaining approaches, suffer from redundant inference.

\begin{figure*}[t]
  \centering
  \subfloat[]{%
  \includegraphics[width=0.7\columnwidth]{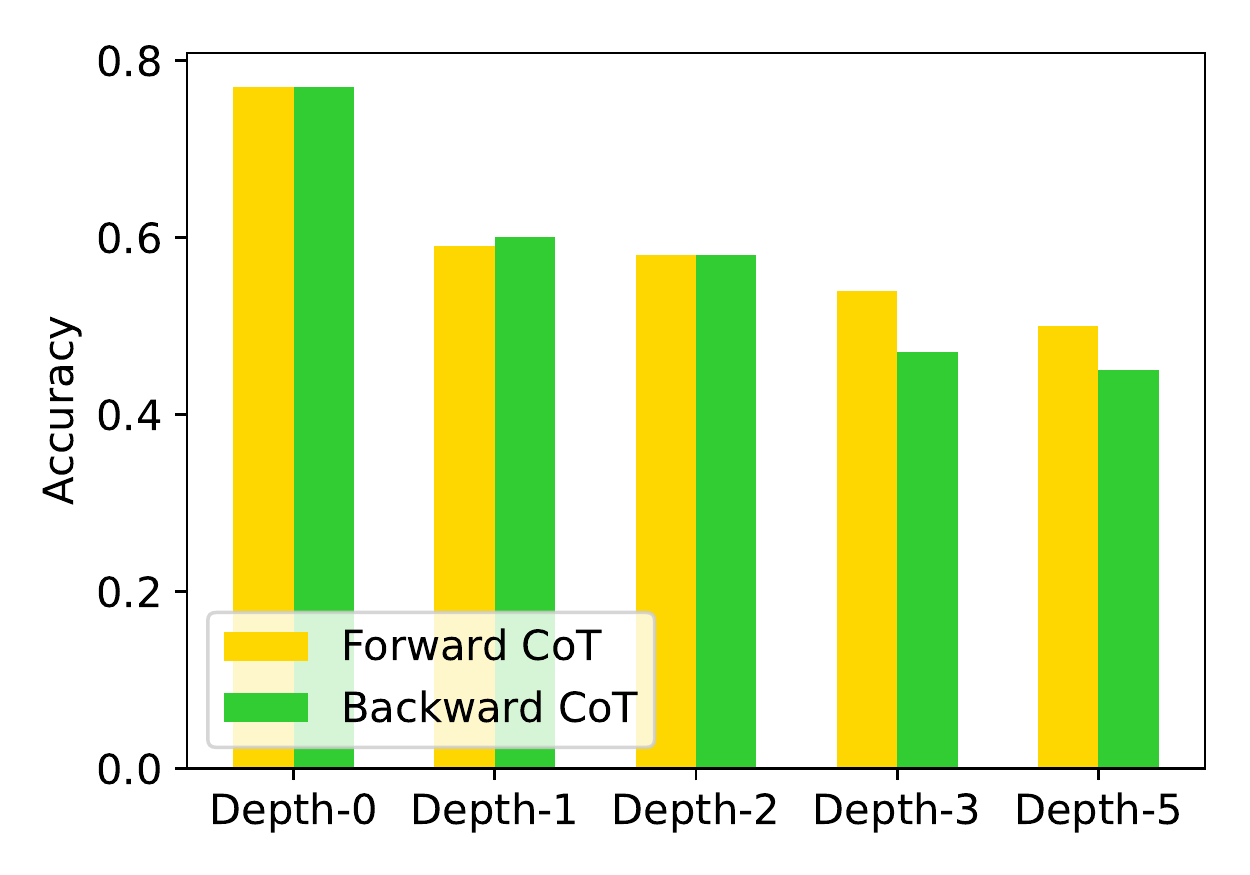}}
  ~~~~\hspace*{0.5cm}
  \subfloat[]{%
  \includegraphics[width=0.7\columnwidth]{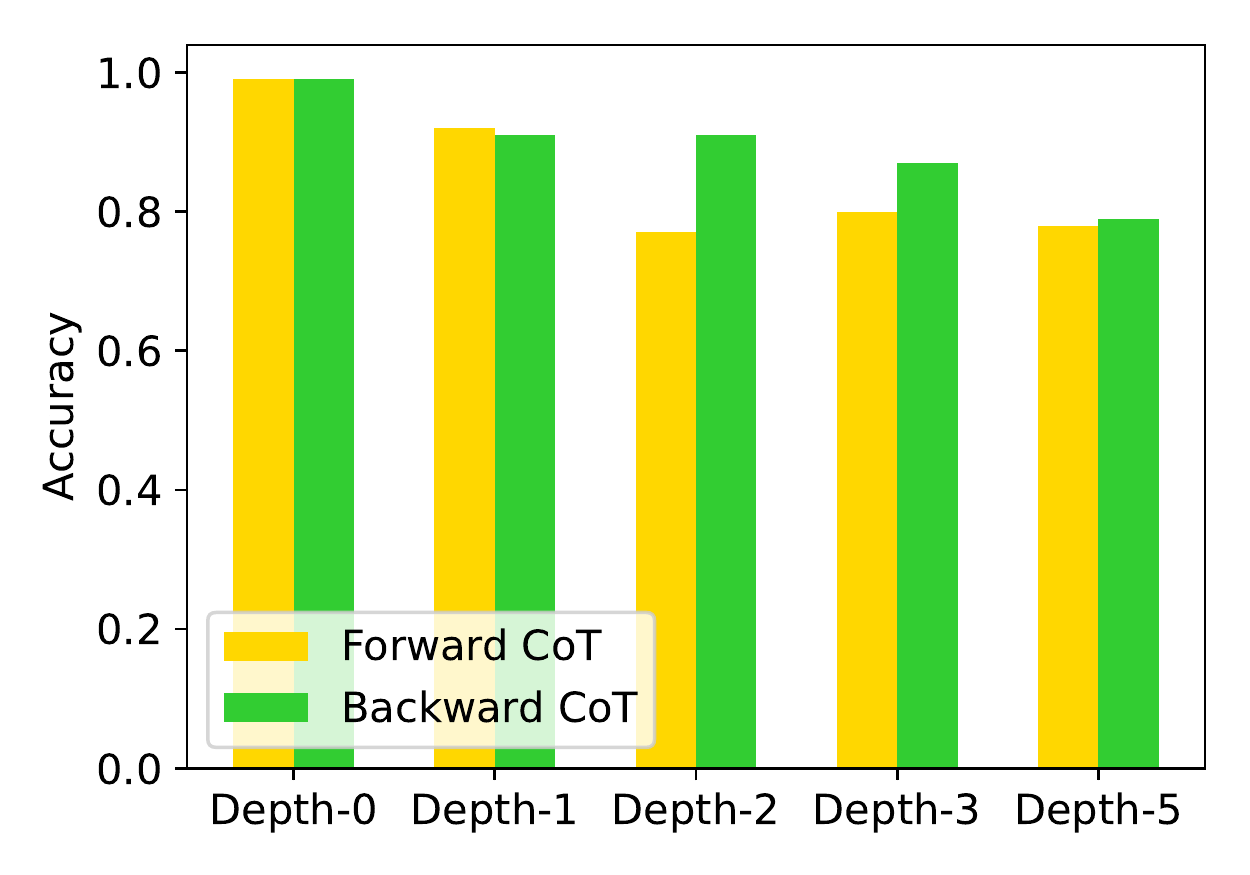}} %
  ~~~~\hspace*{0.5cm}
  \subfloat[]{%
  \includegraphics[width=0.35\columnwidth]{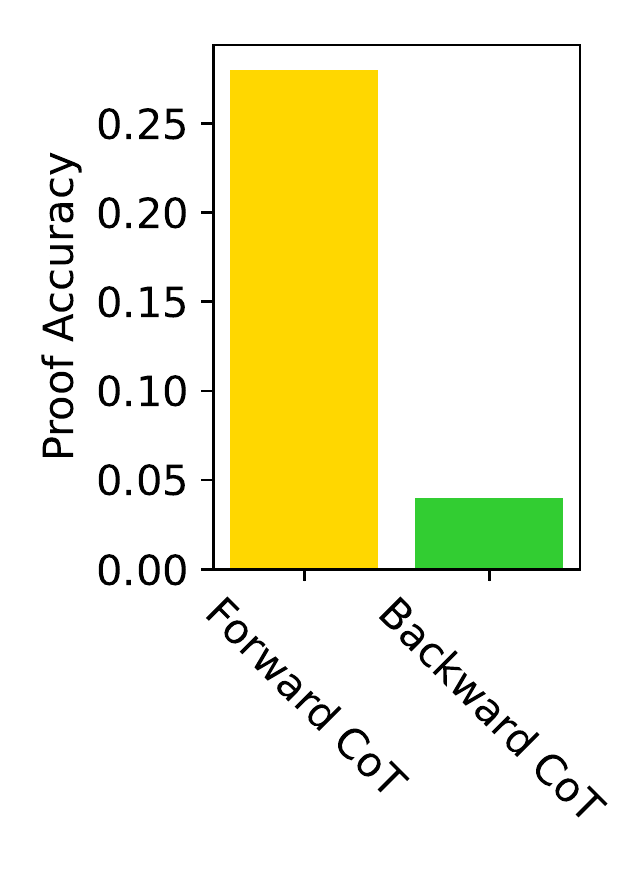}} %

  \caption{%
  \label{fig:backward-cot} %
  Prediction accuracy results on (a) ProofWriter-PUD and (b) ProofWriter-PD with forward and backward CoT. (c) compares the proof accuracy of forward and backward CoT on ProofWriter (Depth-5) for a set of randomly sampled examples for which the models correctly predicted the proof label.}
\end{figure*}

SI also over-predicts \disproved\ in the binary case and \unk\ in the three-way classification case (see Appendix~\ref{sec:confusion-matrix}), performing even worse than the majority class for Depth-5 of PrOntoQA which has more \proved\ labels than \disproved. 

These results, together with Figure~\ref{fig:main-results}, show that backward chaining (which is the backbone of reasoning in \algo) is a better choice compared to forward chaining (the backbone in SI).

\begin{figure}[t]
  \centering
  \includegraphics[width=0.9\columnwidth]{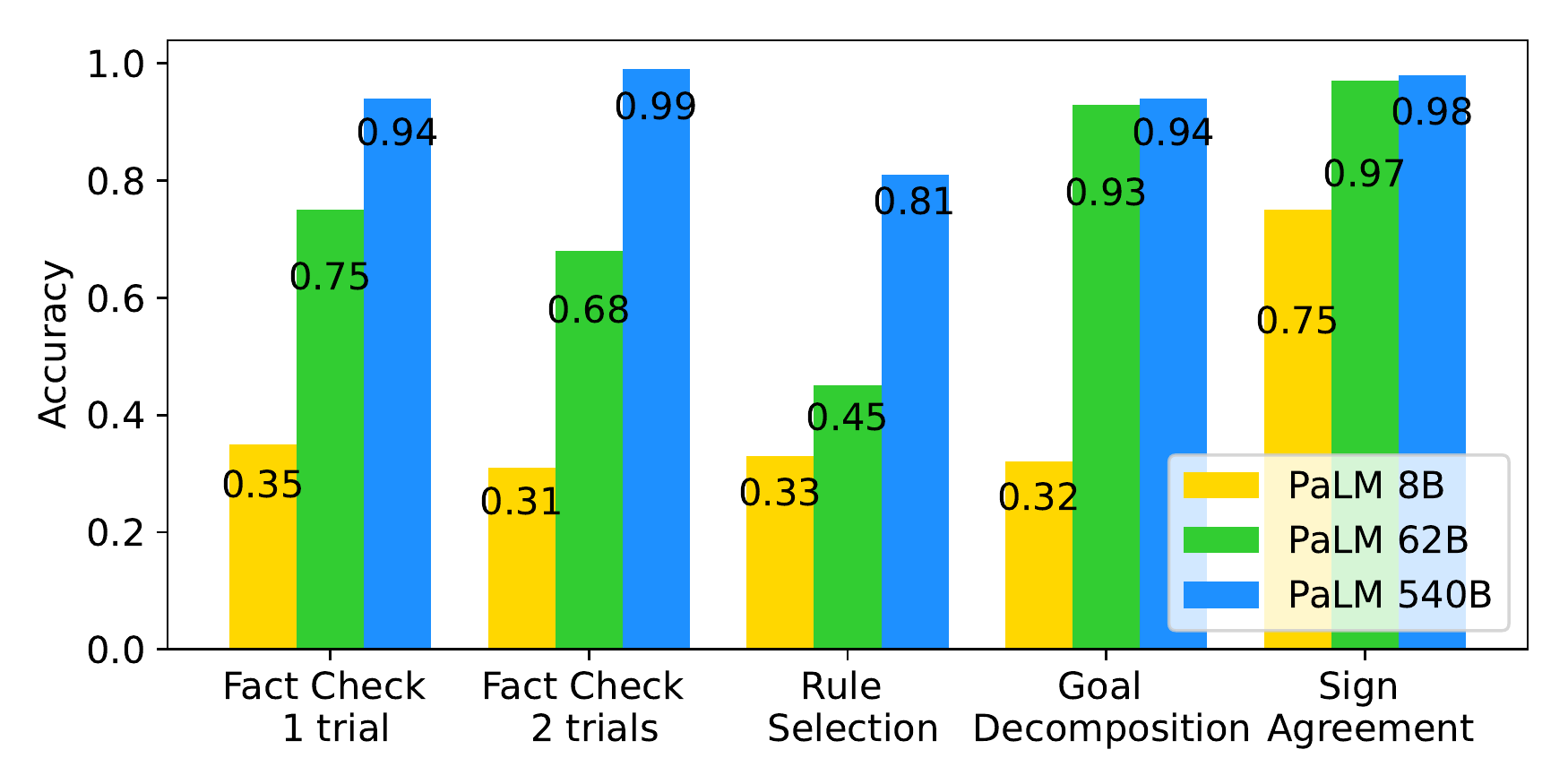}
  \caption{%
  \label{fig:component_analysis} %
    ProofWriter (val) performance of modules in \algo\ in isolation, for different LM sizes.
  }
\end{figure}

\subsection{Does Backward CoT Suffice?}
Our results may raise the question of whether it is enough to directly incorporate the steps of backward chaining into CoT prompts, or if modularity (as in \algo) is also needed. To answer this question, we experiment with a backward version of CoT where the proofs are written in the backward direction from the goal to the premises. The label accuracies are presented in Figure~\ref{fig:backward-cot}(a)--(b) for ProofWriter-PUD and ProofWriter-PD, and their proof accuracy on ProofWriter-PD (Depth-5) in Figure~\ref{fig:backward-cot}(c). The label accuracy of forward and backward CoT are comparable, but forward CoT leads to better performance on PUD and backward CoT leads to better performance on PD. For proof accuracy, however, we see a clear difference between the two versions where backward CoT produces substantially lower quality proofs compared to forward chaining. This result is consistent with the observations of \citet{gontier2020measuring} for finetuned LMs.

The above results show that a modular formulation (as in \algo) is key to successful logical reasoning and simply providing CoT in the backward direction does not suffice. 
We note, however, that future work can use the traces of our model to finetune (smaller) language models (e.g., \citealt{zelikman2022star}), or use the traces as training data in future language models to improve their performance with CoT prompting.

Taking the label and proof accuracy results together, there is also a potential that backward CoT models are more heavily relying on spurious correlations for the PD case where backward CoT outperformed CoT, as backward CoT achieves a similar label accuracy as forward CoT but with a much lower proof accuracy.

\subsection{Qualitative Analysis}
In Figure~\ref{fig:success1}, we show the search trace created by \algo\ for an example from ParaRules, where the answer was predicted correctly. From the figure, one can see how backward chaining helps \algo\ effectively search and create the reasoning chain and how the LM helps fact checking, rule selection, goal decomposition, and sign agreement checking. In Appendix~\ref{sec:appendix-qualitative}, we include an example that has a much larger search trace.

\subsection{Individual Module Analysis}
\label{sec:failure-modes}
To understand which components in \algo\ are responsible for the failure cases, we computed the individual accuracy of the four modules described in Section~\ref{sec:method}. For this purpose, we created four datasets from the validation set of ProofWriter, each measuring only the performance of one module in isolation (see Appendix~\ref{sec:individual-datasets} for details). 

Based on the results of the PaLM 540B model in Figure~\ref{fig:component_analysis}, \module{Rule Selection} is the lowest performing module followed by \module{Goal Decomposition}. 
It is possible that the \module{Rule Selection} module (partially) fails for some examples but \algo\ still arrives at the correct conclusion and proof (e.g., if in Figure~\ref{fig:success1} the third call to \module{Rule Selection} only returned Rule5).
For \module{Fact Check}, when we allow the model to only select one fact, the accuracy is $0.94$ but when we allow the model to select two facts, the accuracy is near perfect. The \module{Sign Agreement} module also shows near-perfect accuracy.

\subsection{The Role of Scale}
We repeat the experiment from Section~\ref{sec:failure-modes} with PaLM 62B and 8B to examine the effect of LM scale on \algo. 
According to the results in Figure~\ref{fig:component_analysis}, when we use PaLM 62B, the performance of the \module{Goal Decomposition} and \module{Sign Agreement} modules remain comparable, but the performance for the \module{Fact Check} and \module{Rule Selection} modules drop substantially. Unlike the first two modules, the second two rely on a one-to-many comparison between the goal and each of the facts/rules which may require a larger model capacity. Moreover, we observe that in PaLM 8B, the accuracy for all components drops significantly, in some cases becoming close to random prediction.

We argue that the extent to which the higher-level reasoning algorithm breaks the problem into sub-problems should be dependent on the scale and power of the base LMs. If smaller LMs are used, then one may need finer-grained problem decomposition (e.g., further decomposing the one-to-many comparisons in the selection module). And as LMs become larger and stronger in the future, one could rely on them to solve problems with a coarser-grained decomposition of the problem.

\begin{figure}[t]
  \centering
  \includegraphics[width=0.85\columnwidth]{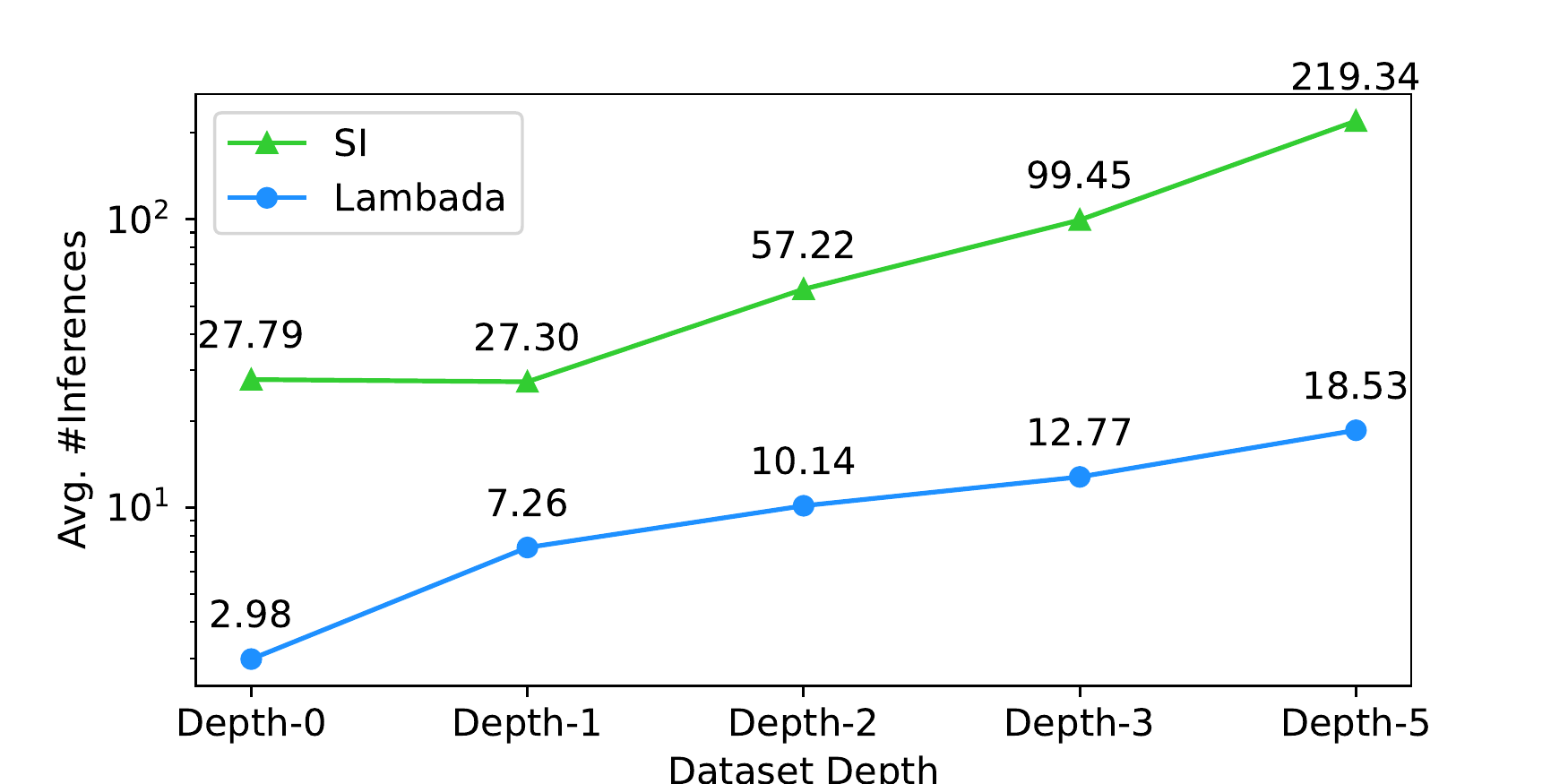}
  \caption{%
  \label{fig:call_count} %
    Comparing \algo\ and SI w.r.t. the average number of inference calls they make per example for different subsets of the ProofWriter-PUD dataset.
  }
\end{figure}

\subsection{Number of Inference Calls}
Another advantage of \algo\ is its efficiency compared to other approaches that require multiple LM inference calls per example such as SI. In Figure~\ref{fig:call_count}, we compare the average number of LM calls per example, for different depths of ProofWriter-PUD. \algo\ requires much fewer calls compared to SI, especially at higher depths: for Depth-1, \algo\ requires 3.8x fewer calls whereas for Depth-5 it requires 11.8x fewer calls.  

\begin{figure}[t]
  \centering
  \includegraphics[width=0.85\columnwidth]{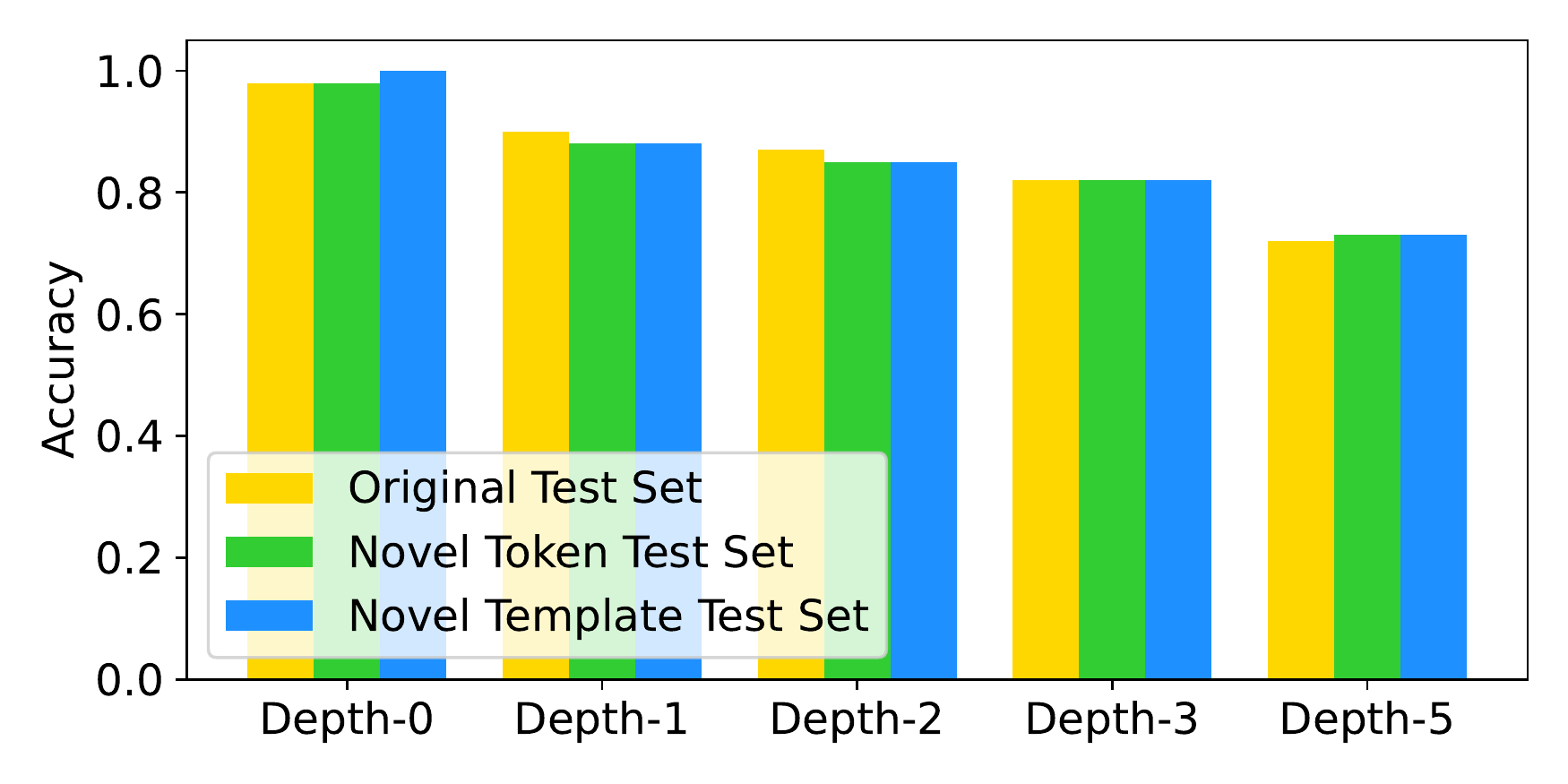}
  \caption{%
  \label{fig:sensitivity-combined} %
    The performance of \algo\ on ProofWriter-PUD for the original, novel token, and novel template test sets.
  }
\end{figure}

\subsection{Lexical Robustness}
To analyze the lexical sensitivity of \algo, we modified the test set of ProofWriter-PUD by replacing various lexical items (names, adjectives, and verbs) with novel tokens and the rule templates with novel ones. We then compared the performance of \algo\ on the original and the modified test sets using the same few-shot examples. The details of the modifications are in Appendix~\ref{sec:sensitivity}. As can be seen in Figure~\ref{fig:sensitivity-combined}, the performance of \algo\ remains almost unchanged, demonstrating robustness to lexical and templatic variations.

\section{Conclusion and Future Directions}
We developed \algo, an algorithm for deductive logical reasoning with natural language that combines the capacity of LMs to handle naturalistic text input with the backward chaining algorithm for robust symbolic reasoning. We showed that \algo\ achieves significant improvements over competitive approaches on challenging benchmarks, both in terms of label accuracy (predicting if a statement can be proved or disproved based on a theory) and proof accuracy. Importantly, this improvement was also observed in a dataset that expresses the theory in more naturalistic expressions, clearly illustrating the benefit of combining an LM with reasoning modules. We also demonstrated the query efficiency and lexical robustness of \algo. 
Although in this paper we only experiment with formal reasoning problems and datasets, we believe our key insight on the efficacy of backward, goal-directed reasoning with LMs has broader implications and can be adapted to other NLP tasks where multi-step inference is required. 

\section*{Limitations}
We identify some limitations and risks with our current work that can be addressed in future work.

\begin{itemize}[nosep,leftmargin=3.5mm]
    \item The current work is mainly applicable to logical entailment problems, where one needs to solve a classification problem of whether a goal can be proved, disproved, or neither proved nor disproved based on a theory. Future work can extend \algo\ to non-classification cases, e.g., where one needs to apply logical reasoning to answer questions such as \texttt{``What color is Fiona?''}.
    \item The current work assumes all the rules are given as input and the rule set is small enough to be included in the prompt. Future work can extend \algo\ to the cases where not all the rules are provided as input and part of the knowledge has to come from the LM itself, as well as the case where not all the rules can be included in the prompt due to the limitation in the prompt length.
    \item The current work is limited to deductive reasoning with the \emph{modus ponens} rule; future work can expand the applicability of \algo\ on datasets with other types of rules such as proof by contradiction, disjunction elimination, etc.
    \item The calls made to the LM modules in \algo\ are dependent on the value from the previous call. That is, we need to wait for the results from one call before we decide what the next call must be. Since making batch calls to the LMs is typically easier and faster, future work can find ways to implement \algo\ with batch LM calls.
    \item While we showed that \algo\ is more efficient than SI in terms of the number of inference calls it makes to the LM, it still requires many more calls to the LM compared to approaches such as CoT, hence increasing the required computation and cost.
\end{itemize}

\bibliography{bib}
\bibliographystyle{acl_natbib}

\appendix

\begin{figure*}[t]
  \centering
  \includegraphics[width=\textwidth]{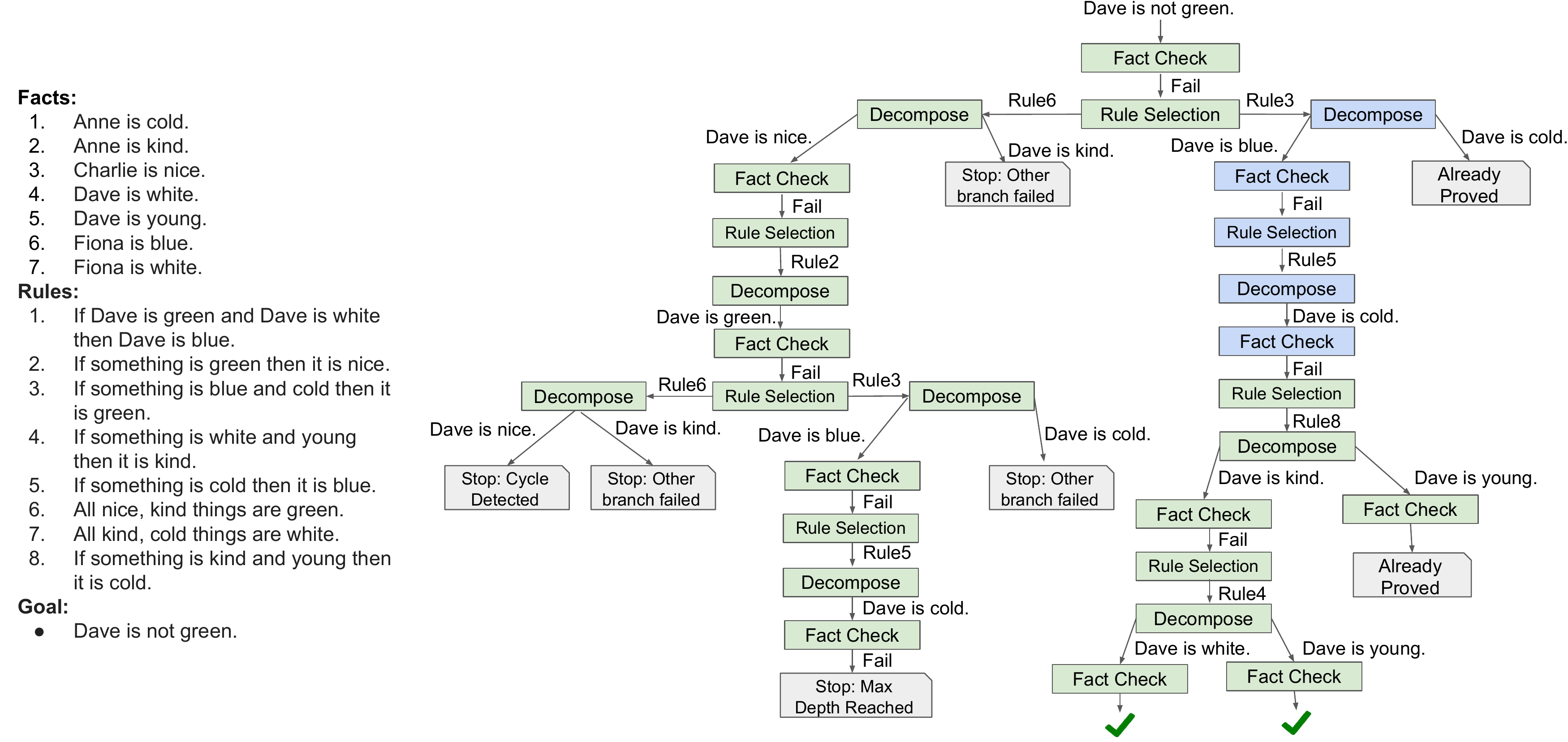}
  \caption{%
  \label{fig:success-depth5} %
    The search trace of \algo\ on an example from ProofWriter with depth=5 where the answer was predicted correctly. The sign agreement module has been omitted for brevity. The modules color-coded with blue represent the calls where the module retrieved the value from the cache instead of calling the LM.
  }
\end{figure*}

\section{Caching and Avoiding Loops for \algo} \label{sec:cache}
Since \algo\ is a recursive algorithm, during the proof of an example Algorithm~\ref{algo:back-chain} may be called with the same goal multiple times. For instance, consider the goal \texttt{``Eric is nice''} for the theory in Figure~\ref{fig:success1}. Applying Rule6 breaks the goal into three sub-goals. The first one is \texttt{``Eric is big''} which is proved using the \module{Fact Check} module. For the second sub-goal, Rule3 is used to compose it into three sub-goals the first of which we have proved before. 
Since we have already proved this sub-goal, we can save a \module{Fact Check} call if we cache previous results.

Note that the result of a call to \algo\ can be different depending on the input max depth. For example, the algorithm may return \unk\ when called for the theory and goal in Figure~\ref{fig:success1} with max depth 0, and return \proved\ when called with max depth 3. Specifically, if we can prove/disprove a goal at depth $d$, we can conclude that it can be proved/disproved at depths $\geq d$ as well and we can get the value from the cache. Moreover, if the algorithm returns \unk\ for a goal at depth $d$, we can conclude that it will also return \unk\ at depths $<d$. Therefore, if the algorithm is called for a theory and goal at depth $d$, we also check other depths to see if we have the results for other depths that apply to this case. 
Besides having a cache for the entire algorithm that avoids redundant computations when the truth of a goal has been previously computed for a theory, each individual module can also have its own cache as it is possible that the module is called for the same theory and goal. We show one such example in Figure~\ref{fig:success-depth5} (to be discussed in Section~\ref{sec:more-results}).

\algo\ may sometimes run into loops. For example, to prove a (sub-)goal \texttt{``Fiona is round?''}, after recursively identifying rules that unify with it and decomposing it into sub-goals, the algorithm may arrive at a point where it needs to prove the \texttt{``Fiona is round?''} sub-goal, which is equivalent to the initial goal. To avoid such loops, for each path in the proof trace, we keep track of the (sub-)goals that are to be proved and stop further exploring that branch of the search trace when a loop is identified.

Note that in Algorithm~\ref{algo:back-chain}, for clarity of the algorithm we did not include the caching and loop avoidance operations. Also note that caching and loop avoidance mainly help with reducing the number of inference calls.

\begin{figure*}[t]
  \centering
  \includegraphics[width=\textwidth]{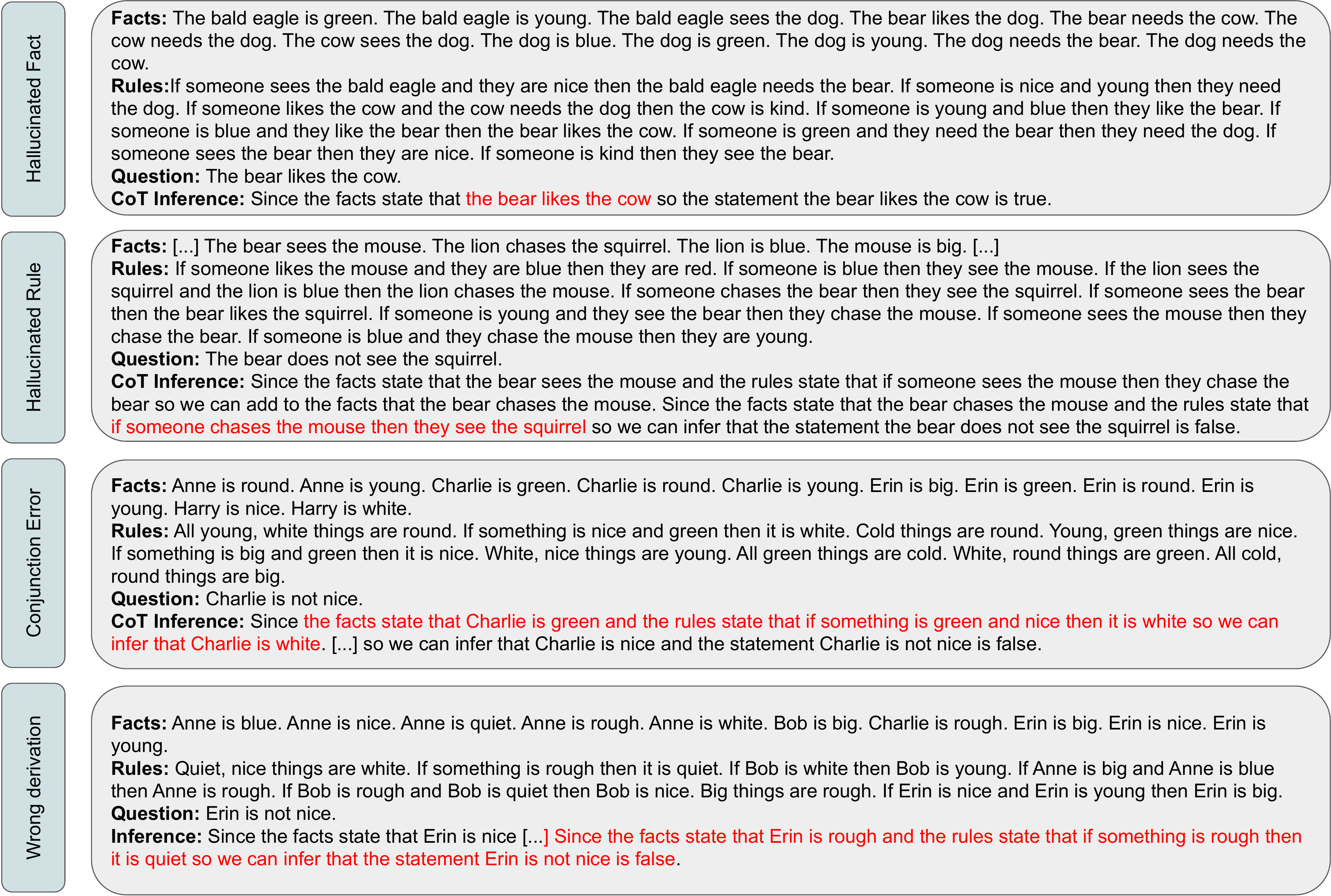}
  \caption{%
  \label{fig:cot-failures} %
    Examples of wrong CoT proof chains from four different categories. The erroneous part is marked in red. 
  }
\end{figure*}

\section{Additional Results and Analyses} \label{sec:more-results}
In this section, we provide some more in-depth qualitative and quantitative analysis of the results from our model and the baselines. 

\subsection{Qualitative Analysis} \label{sec:appendix-qualitative}
In Figure~\ref{fig:success-depth5}, we provide the search trace of \algo\ for an example in ProofWriter (Depth-5) for which \algo\ correctly predicted that the goal is disproved based on the theory. We deliberately selected an example with a large search trace to demonstrate the various aspects of \algo.

\algo\ starts by calling the \module{Fact Check} module on the goal which fails to prove or disprove it. So \module{Rule Selection} is called which identifies two rules that can be applied: Rule3 and Rule6. Since Rule6 is shorter, the reranker ranks it higher; \algo\ starts with this rule and calls the \module{Goal Decomposition} module which breaks the goal into two sub-goals: \texttt{``Dave is nice.''} and \texttt{``Dave is kind.''}. Starting with the first sub-goal, \module{Face Check} fails on it so \module{Rule Selection} is called which selects Rule2 and \module{Goal Decomposition} decomposes the sub-goal into \texttt{``Dave is green.''}. 

Note that if the cycle checking was smart enough to understand that this sub-goal is the negation of the root goal, we could stop further searching this branch. However, we currently only do cycle matching for exact matches so the algorithm continues the search trace.

\module{Fact Check} fails again so \module{Rule Selection} is called which selects Rule3 and Rule6 again, and since Rule6 is shorter the algorithm continues with that rule. \module{Goal Decomposition} breaks the sub-goal into \texttt{``Dave is nice.''} and \texttt{``Dave is kind.''}. Considering the first sub-goal, the algorithm identifies a cycle and stops the search. The second sub-goal is also ignored as there is a conjunction between the sub-goals.

The algorithm then continues with calling \module{Goal Decomposition} for Rule3 which breaks the sub-goal into \texttt{``Dave is blue.''} and \texttt{``Dave is cold.''}. Starting with the first sub-goal, since \module{Fact Check} fails the algorithm calls \module{Rule Selection} which selects Rule5 and \module{Goal Decomposition} breaks the sub-goal into \texttt{``Dave is cold.''}. \module{Face Check} fails on this sub-goal and since the maximum depth is reached, the algorithm stops expanding this branch. Moreover, the branch for \texttt{``Dave is cold.''} is no longer pursued because there was a conjunction between the sub-goals and one of them failed.

Moving on to the right branch in Figure~\ref{fig:success-depth5}, the algorithm calls the \module{Goal Decomposition} module for the goal and Rule3. Since we have previously computed it, the sub-goals \texttt{``Dave is blue.''} and \texttt{``Dave is cold.''} are returned from the cache. \module{Fact Check} is called on \texttt{``Dave is blue.''} and since it has been computed before, the result (failure) is retrieved from the cache. The \module{Rule Selection} module is called, where the result (Rule5) is again retrieved from the cache. \module{Goal Decomposition} is then called and the sub-goal \texttt{``Dave is cold.''} is retrieved from the cache. \module{Fact Check} fails again (retrieved from the cache), \module{Rule Selection} selects Rule8 and \module{Goal Decomposition} produces two sub-goals: \texttt{``Dave is kind.''} and \texttt{``Dave is young.''}. For \texttt{``Dave is kind.''}, \module{Fact Check} fails, \module{Rule Selection} selects Rule4 and \module{Goal Decomposition} produces two sub-goals: \texttt{``Dave is white.''} and \texttt{``Dave is young.''}. For both of these sub-goals, \module{Fact Check} succeeds in proving them. The algorithm then also checks \texttt{``Dave is young.''} for the right branch, but since this sub-goal has already been proved, it just gets the result from the cache.  The algorithm then checks \texttt{``Dave is cold.''} for the rightmost branch, but since this sub-goal has already been proved, it just gets the result from the cache.

The model also calls the \module{Sign Agreement} module for rules on the right branch (not shown in the Figure) and finds out that the sign of the rules and the sub-goals agree for all cases, except for the very first rule selected (Rule3) so it correctly concludes that the goal is disproved.

\begin{figure*}[th!]
  \centering
  \subfloat[ProofWriter-PUD (Depth-5)]{%
  \includegraphics[width=0.3\textwidth]{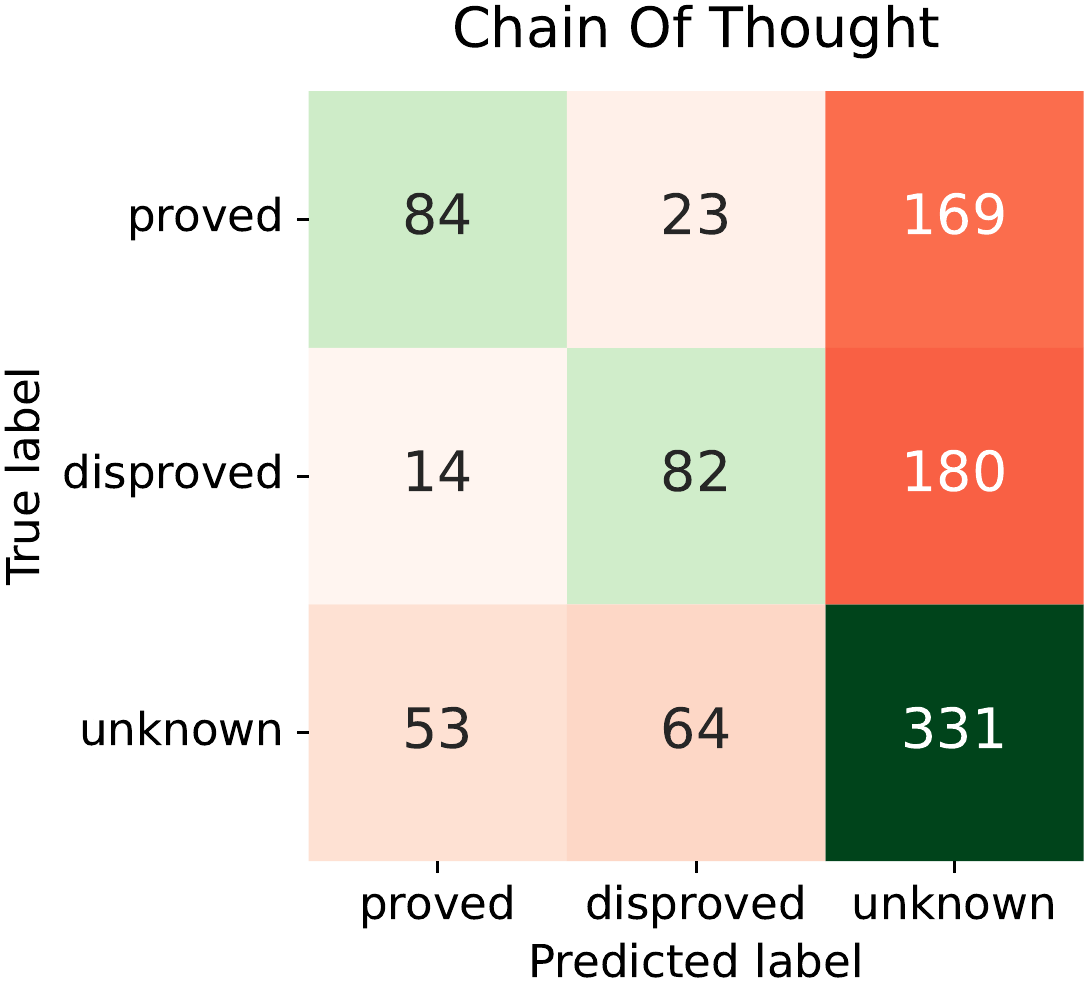}}
~~~~\hspace*{0cm}
  \subfloat[ProofWriter-PUD (Depth-5)]{%
  \includegraphics[width=0.3\textwidth]{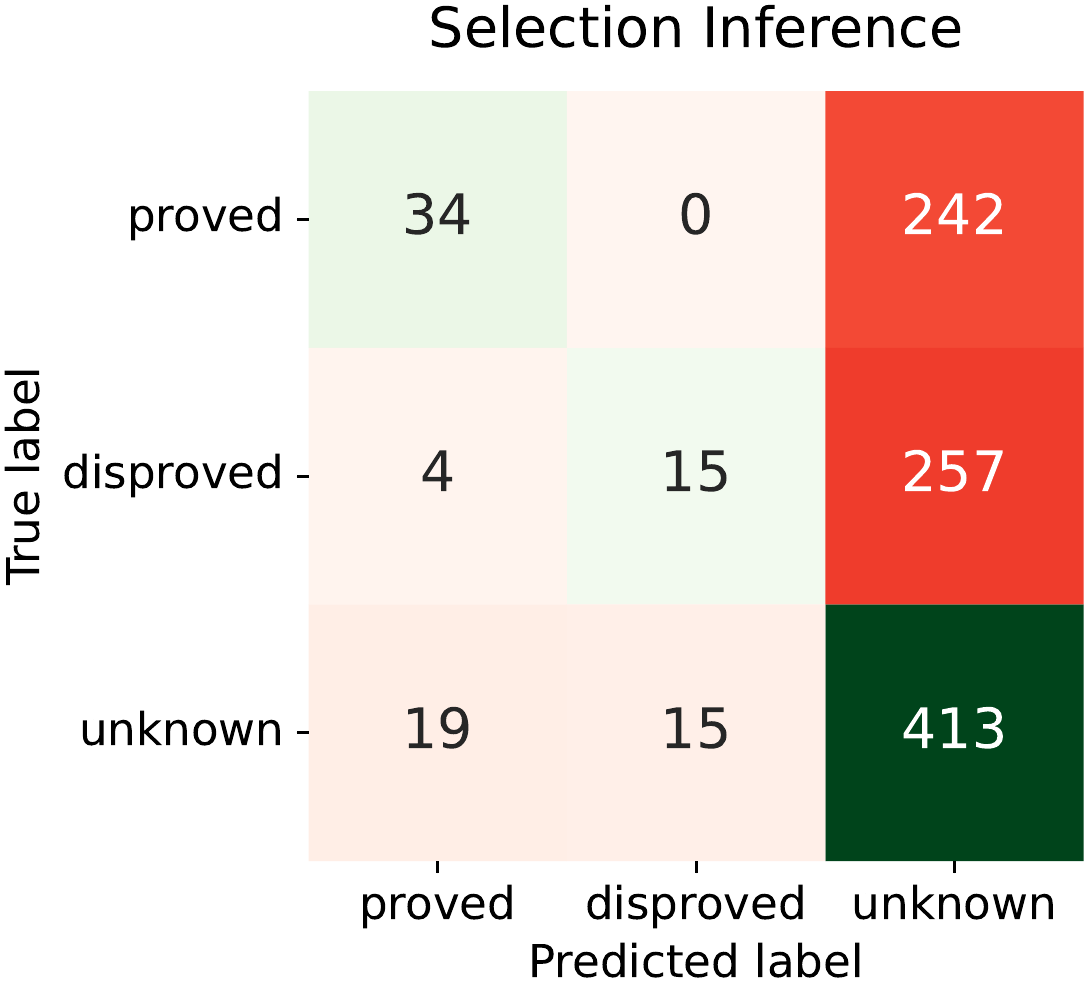}} %
~~~~\hspace*{0cm}
    \subfloat[ProofWriter-PUD (Depth-5)]{%
  \includegraphics[width=0.3\textwidth]{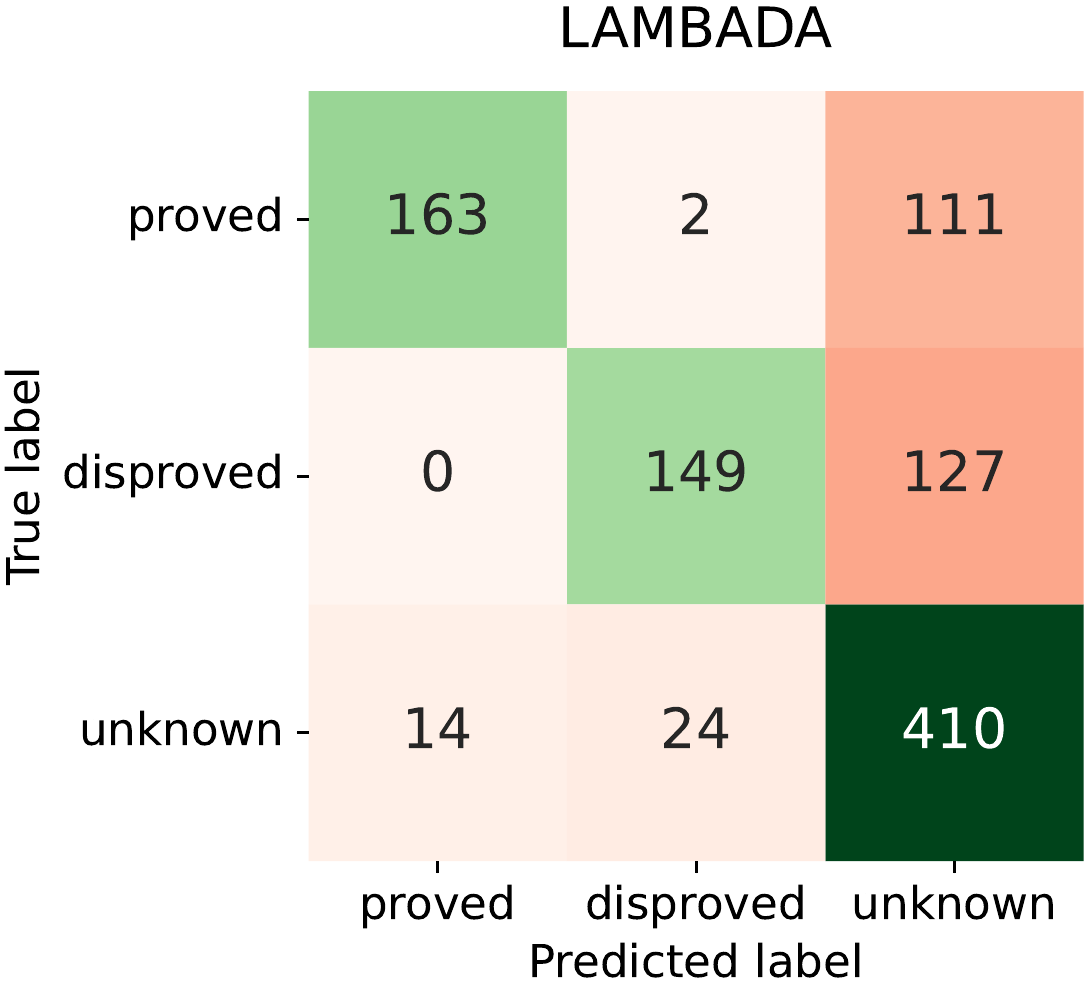}} %
  \\
  \subfloat[PrOntoQA (Depth-5)]{%
  \includegraphics[width=0.3\textwidth]{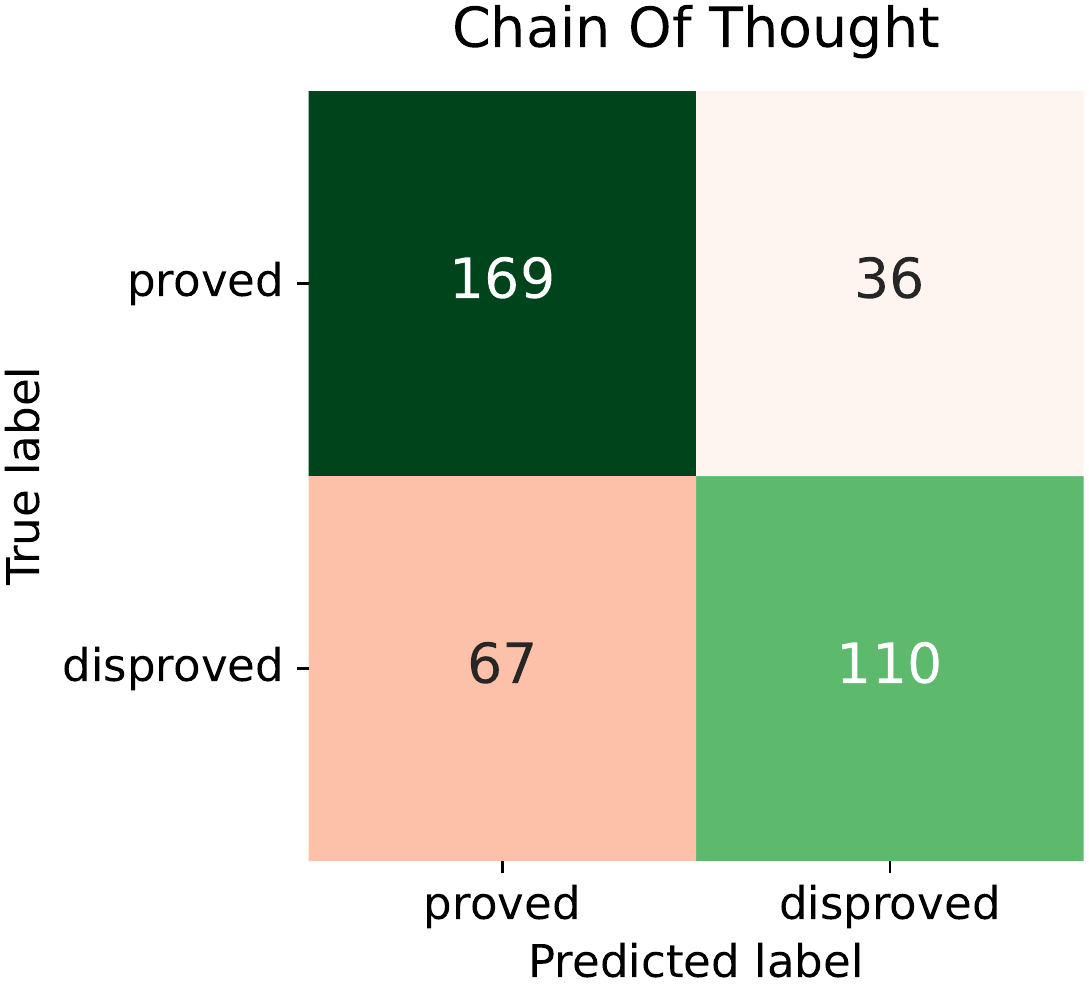}}
~~~~\hspace*{0cm}
  \subfloat[PrOntoQA (Depth-5)]{%
  \includegraphics[width=0.3\textwidth]{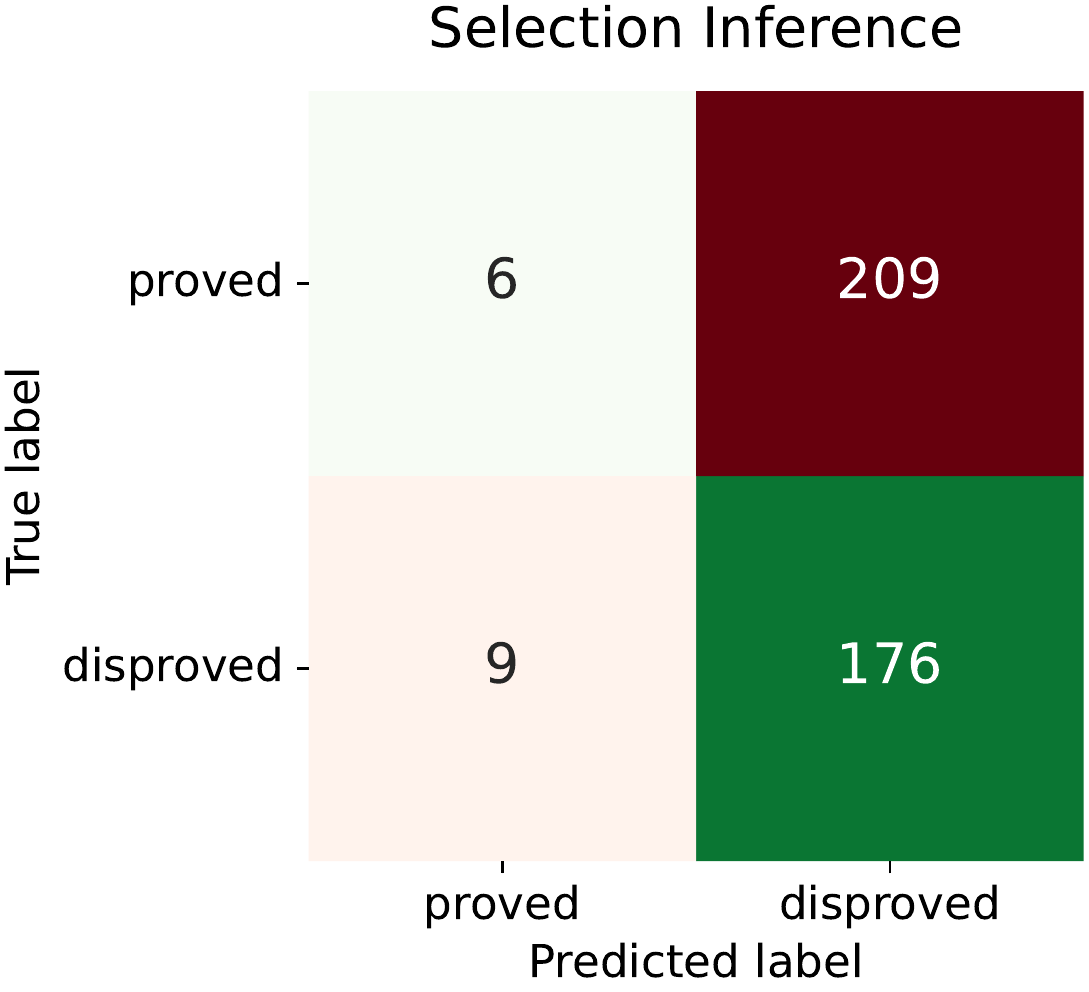}} %
~~~~\hspace*{0cm}
    \subfloat[PrOntoQA (Depth-5)]{%
  \includegraphics[width=0.3\textwidth]{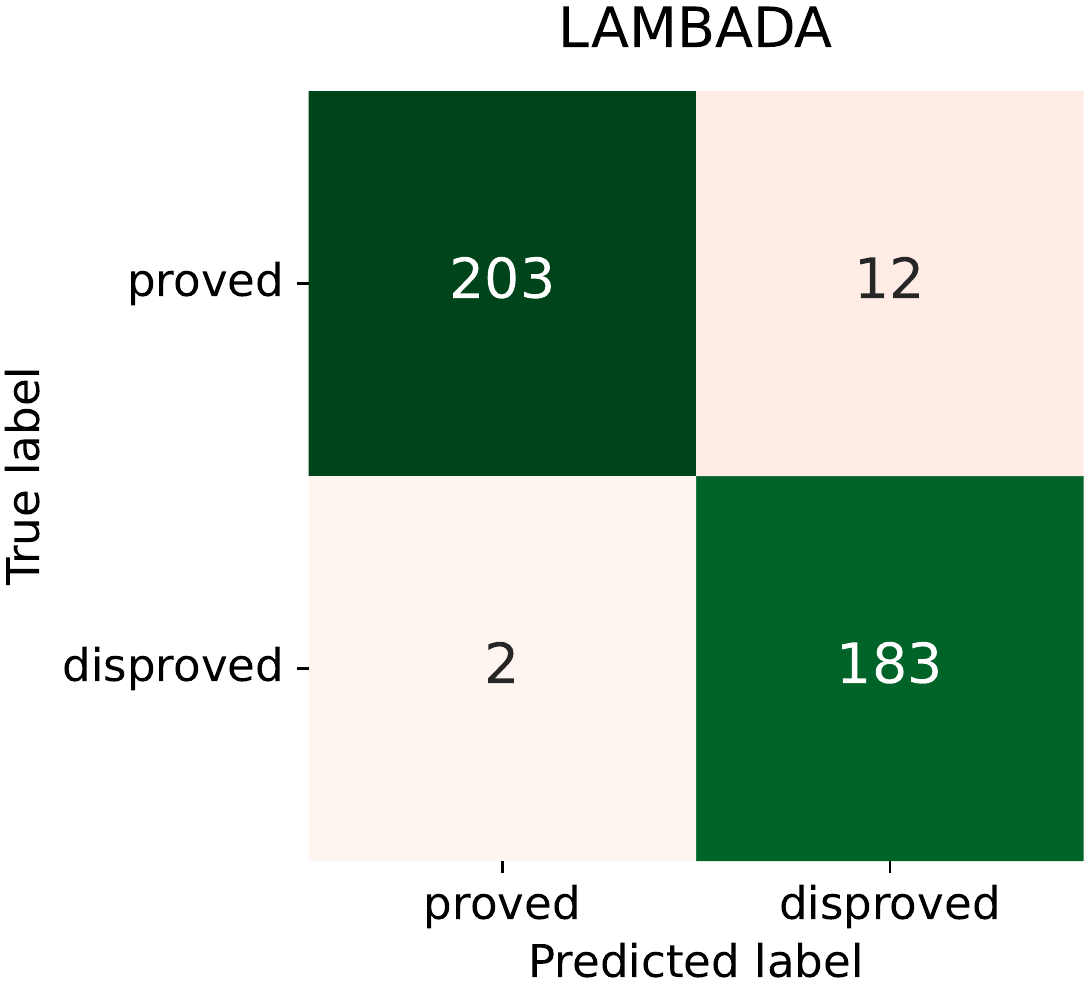}} %
  \\
  \subfloat[ParaRules]{%
  \includegraphics[width=0.3\textwidth]{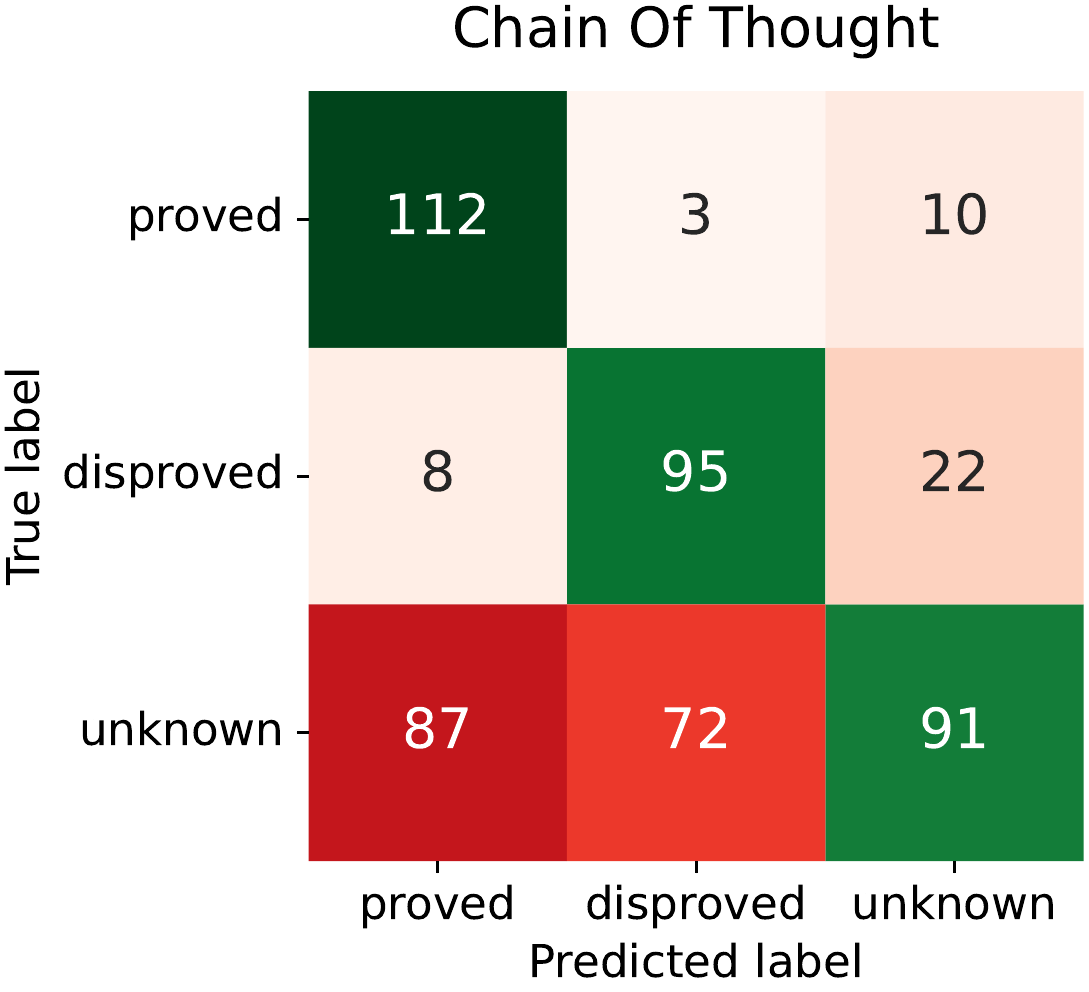}}
~~~~\hspace*{0cm}
  \subfloat[ParaRules]{%
  \includegraphics[width=0.3\textwidth]{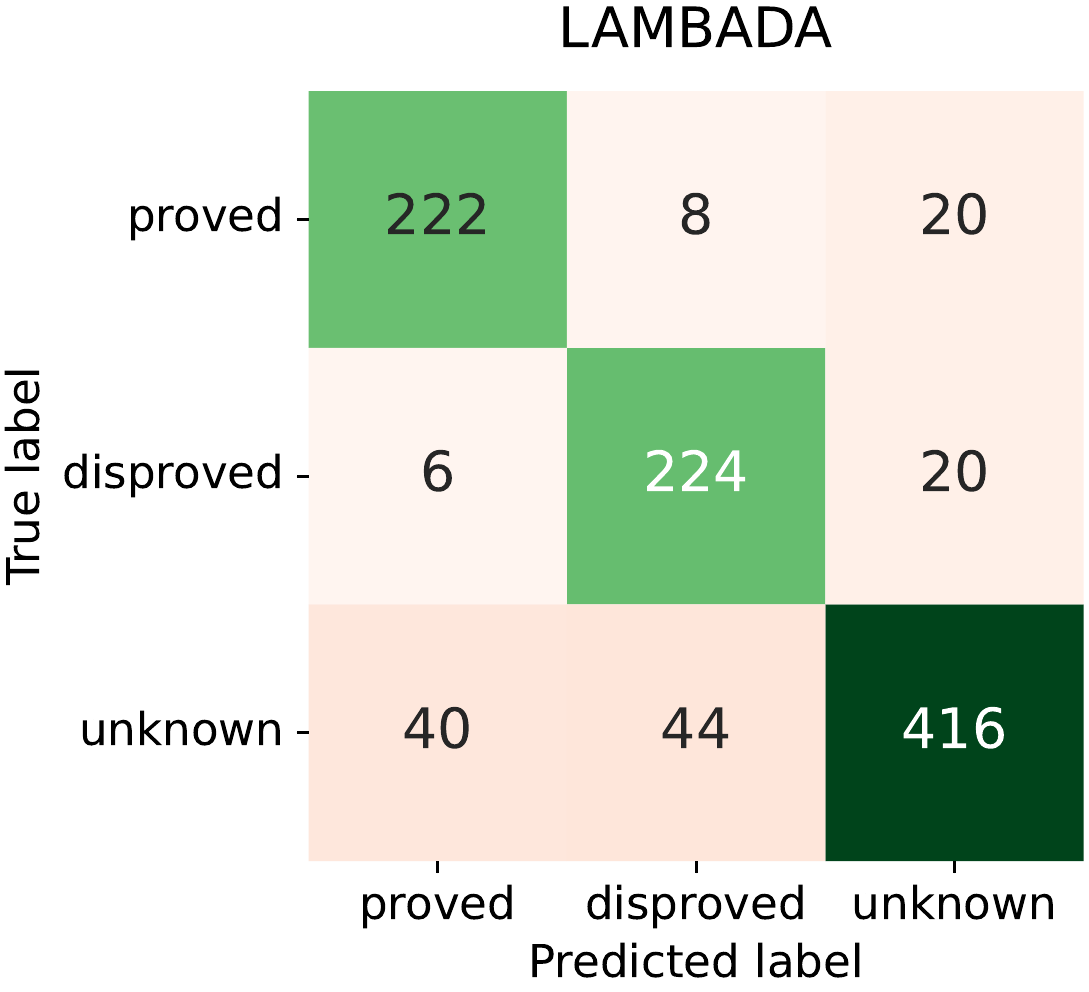}} %
  \caption{%
  \label{fig:confusion_matrices} %
  Confusion matrices.}
\end{figure*}

\subsection{Further Analysis of CoT} \label{sec:cot-proof-errors}
In Figure~\ref{fig:main-results}(e), we observed that CoT mostly produces wrong proof chains even when the predicted label is correct. Through manually analyzing $50$ examples for which CoT predicted the correct label, we identified three dominant reasons for the chains being wrong: 1- hallucinating rules or facts, 2- not understanding conjunction, and 3- making invalid derivations. In Figure~\ref{fig:cot-failures}, we show failure examples from each category. Notice that, e.g., in the example with a hallucinated rule, CoT relies on a rule \texttt{``if someone chases the mouse then they see the squirrel''} which not only does not appear in the provided set of rules, but cannot even be derived with a combination of the rules.

The high label accuracy of CoT and its low proof accuracy on ProofWriter-PD hint at the possibility of spurious biases that can be exploited by CoT.
For example, we found that in $9.2\%$ of the examples which require 1+ reasoning hops, the consequent of one of the rules in the theory is the same as the goal to be proved, and for $98.9\%$ of these examples the label is \proved. In several of these examples, CoT simply concluded that the goal can be proved in $0$ hops based on a hallucinated fact. Moreover, the existence of the word \texttt{``not''} in the goal is highly predictive of the label: goals having \texttt{``not''} are mostly \disproved\ and goals not having \texttt{``not''} are mostly \proved. The PUD case solves the latter issue to a large extent as the label for a good portion of the examples with or without \texttt{``not''} in \unk.
The spurious correlations also explain the fluctuations in the CoT performance across different depths, as the performance depends on how much those correlations appear in the few-shot demonstrations.

We reiterate that for SI and \algo, such spurious correlations between the input and the label cannot be exploited because the intermediate modules are impervious to the correlations between the input and the label.

\subsection{Forward Chaining Becomes Progressively More Difficult} \label{sec:si-becomes-harder}
Algorithms such as SI that are based on forward chaining require a combinatorial search of the theory to find the right subset of facts and rules in each step of the reasoning. The search space becomes progressively larger as the algorithm makes new inferences and those inferences are added back to the theory. For example, if the initial size of the theory (i.e. the number of facts plus the number of rules) is $|\rules{C}|$, when making the $k$-th inference the size of the theory is $|\rules{C}|+k-1$. 

Conceptually, as the model produces more inferences, the distance to the goal (in terms of the number of hops remaining between the goal and the facts) should reduce and so the later inferences should be more accurate. However,
we hypothesize that the increase in the size of the theory (and hence the size of the search space) may result in lower success rates in the later inferences of the SI model. 
To verify this experimentally, we further analyzed the results of SI on depth-5 of PrOntoQA as follows. We extracted the subset of examples where the label was \proved\ but SI failed to find a proof (these are examples where at least one of the inferences is not on the proof chain). Then, as a proxy for measuring the responsibility of the $k$-th inference of the model for the failure, we measured the percentage of times the $k$-th inference was on the proof chain (the proof chain for each test example is provided as part of the dataset). Notice that it is possible that, e.g., the first inference is not on the proof chain, but the rest of the inferences are. The results are reported in Figure~\ref{fig:si-becomes-harder} in the main text. 
The results show that the chance of producing inferences that are on the proof chain progressively decreases in the later inferences of the model where the size of the input theory (and hence the search space) is larger.

\subsection{Confusion Matrices} \label{sec:confusion-matrix}
We reported the overall model accuracies in the main text. Here, we report finer-grained confusion matrices that help better understand the biases of the model. Figure~\ref{fig:confusion_matrices} reports the confusion matrices for our datasets. According to the results, we observe that whenever \algo\ predicts \proved\ or \disproved, the prediction is mostly correct. The accuracy is slightly more on cases where the prediction is \proved\ than \disproved. We believe this is because \disproved\ cases typically involve negation that makes the reasoning more complex. However, there are several examples for which the label is \proved\ or \disproved, whereas the model predicts \unk.

CoT and SI also show similar behaviour as \algo\ on ProofWriter-PUD but with a larger bias toward prediction \unk. Moreover, SI shows a large tendency toward predicting \disproved\ for PrOntoQA.

\begin{figure*}[t]
  \centering
  \subfloat[]{%
  \includegraphics[width=0.45\textwidth]{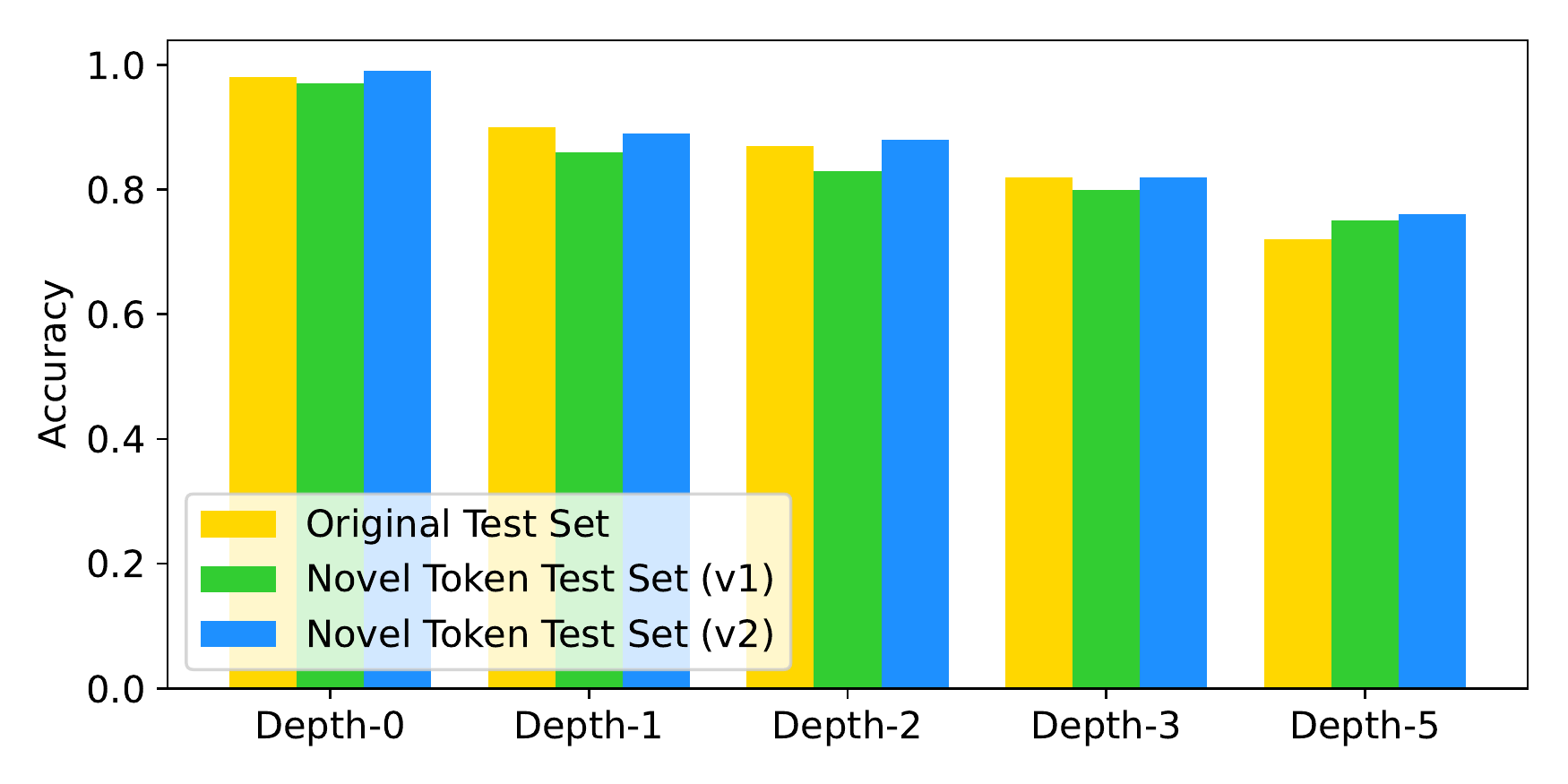}}
~~~~\hspace*{0cm}
  \subfloat[]{%
  \includegraphics[width=0.45\textwidth]{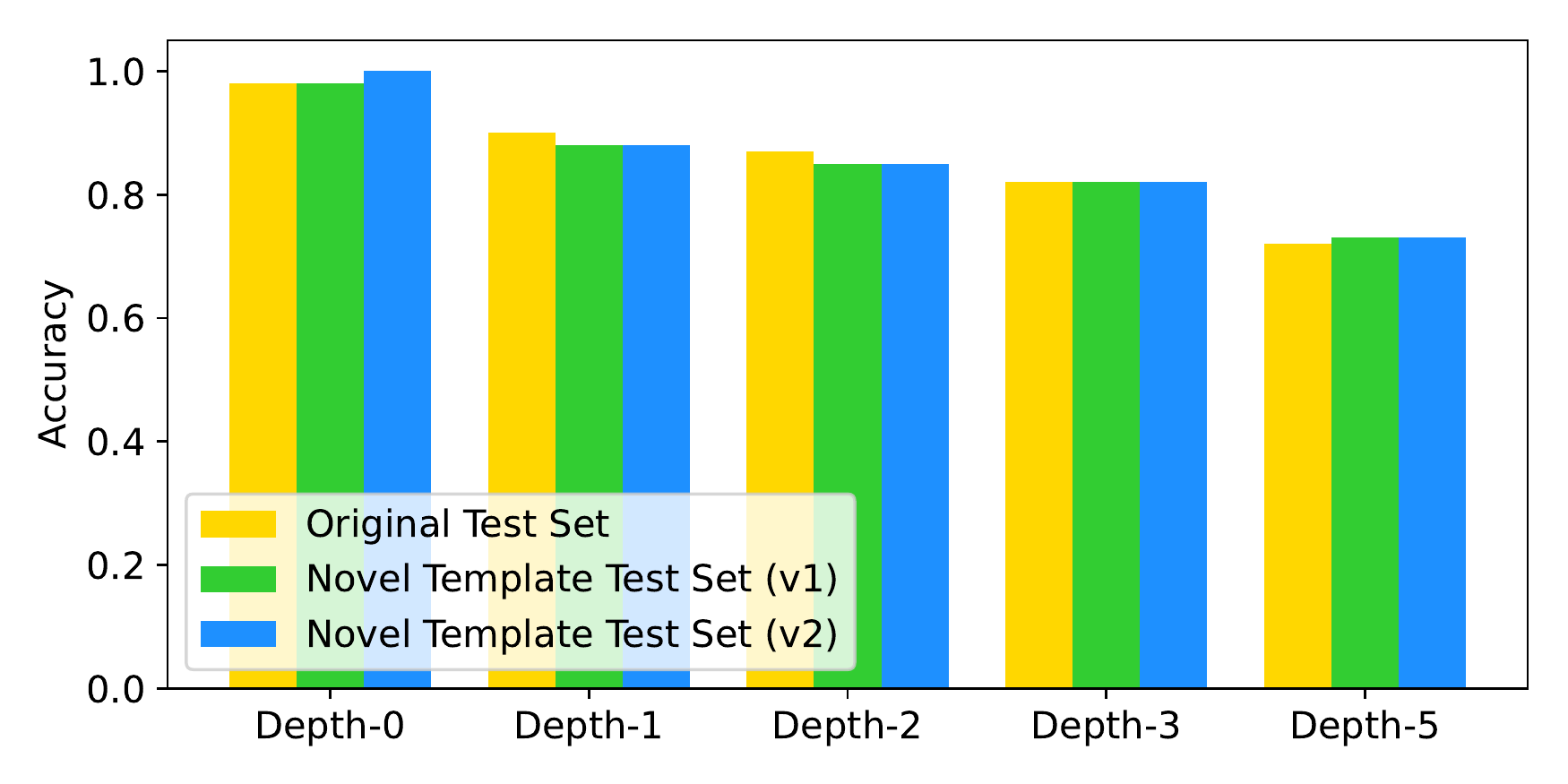}} %
  \caption{%
  \label{fig:sensitivity} %
    The performance of \algo\ on ProofWriter-PUD for (a) the original and the novel token test sets, (b) the original and the novel template test sets. The results show that \algo\ is robust to lexical and template modifications.
  }
\end{figure*}

\subsection{Lexical Sensitivity Analysis} \label{sec:sensitivity}
To analyze the lexical sensitivity of \algo, we created a new test for ProofWriter-PUD which contains tokens that do not appear in demonstration examples. Specifically, we manually created a pool of entity names, animal names, adjectives, and verbs (all of them previously not appearing in the ProofWriter dataset) and then made the following modifications for each example: 1- identified all entity names and mapped each entity name to a randomly selected name from the pool, 2- identified all animals and mapped each of them to a randomly selected animal from the pool, 3- identified all adjectives and mapped each of them to a randomly selected adjective from the pool, and 4- identified all verbs and mapped each of them (except the \emph{to be} verbs) to a randomly selected verb from the pool. As an example, \texttt{dog} may be mapped to \texttt{bison} in one example and to \texttt{camel} in another. Then, using the same few-shot examples as before, we tested the performance of \algo\ on this modified test set and compared the results to the original test set.

We also analyzed the sensitivity to the templates used for the rules. Toward this goal, we identified the templates used for the rules in the ProofWriter dataset and replaced each template with another template (previously not appearing in the ProofWriter dataset). For example, we changed the template \texttt{``[X] things are [Y]''} to \texttt{``It is a truth that [X] things are always [Y] as well''}. Then, using the same few-shot examples as before, we tested the performance of \algo\ on this modified test set and compared the results to the original test set.

We repeated the aforementioned experiments twice for each analysis each time using a different set of tokens/templates. The results in Figure~\ref{fig:sensitivity-combined} in the main text demonstrate the average accuracy across two runs. The results for individual runs are presented in Figure~\ref{fig:sensitivity}(a),~(b) for the two analyses respectively. According to the results, while we observe some variations in the total accuracy (for some depths the performance goes slightly down and for some depths goes slightly up), the performance stays in the same ballpark, showing the robustness of \algo. Moreover, comparing the results on the modified test set with those of the baselines reported in the main text, we observe that even on this modified test set, \algo\ performs significantly better than the baselines tested on the original test set.

\section{Combinatorial Search Issue in Forward Chaining} \label{sec:combinatorial-search}
Consider a simple fictional theory with the following facts:\\
$[$\texttt{Anne is cold.}, \texttt{Anne is nice and pink.}, \texttt{Anne is kind.}, \texttt{Anne is green.}, \texttt{Anne is big and young.}, \texttt{Anne is rough.}, \texttt{Anne is round.}$]$\\
the following rules:\\
$[$\texttt{Cold, red people are white.}, \texttt{Nice, blue people are white.}, \texttt{Kind, green people are white.}, \texttt{Cold, round people are white.}, \texttt{Big, green people are white.}$]$\\
and the goal \texttt{``Anne is white.''}. An approach based on forward chaining requires selecting a subset of the facts and rules from the theory from which this goal can be proved. Specifically, it needs to select \texttt{``Anne is cold.''}, \texttt{``Anne is round.''}, and \texttt{Cold, round people are white.} from the theory.
Such a selection requires a combinatorial search where different combinations of facts and rules should be tested to see which one can lead to proving the goal. An LM may fail to search this space effectively in a single inference call. 

SI uses an approximation to reduce the search space: it first makes an inference call to an LM to select one fact/rule, then it makes another inference call to select the next fact/rule based on the first one, and continues to make inference calls until a halting criterion is met. This approximation reduces the search space from a combinatorial space to a linear space. Since the facts/rules are not selected jointly, however, the chances of selecting the wrong combinations of facts and rules increase because repairing a wrong first choice is not possible, and this leads to low performance as evidenced in our experimental results.

With a backward chaining approach such as \algo, on the other hand, no combinatorial search (or approximations to it) is required: the \module{Rule Selection} module verifies each rule independently to see which one is applicable (i.e. a linear scan), the \module{Goal Decomposition} module breaks goals into sub-goals based on each selected rule independently of the other selected rules, and the \module{Fact Check} module verifies the existence of a fact that entails or contradicts the goal with a linear search over the facts.

\section{Implementation Details} \label{sec:impl_details}
\begin{figure}[t]
  \centering
  \includegraphics[width=\columnwidth]{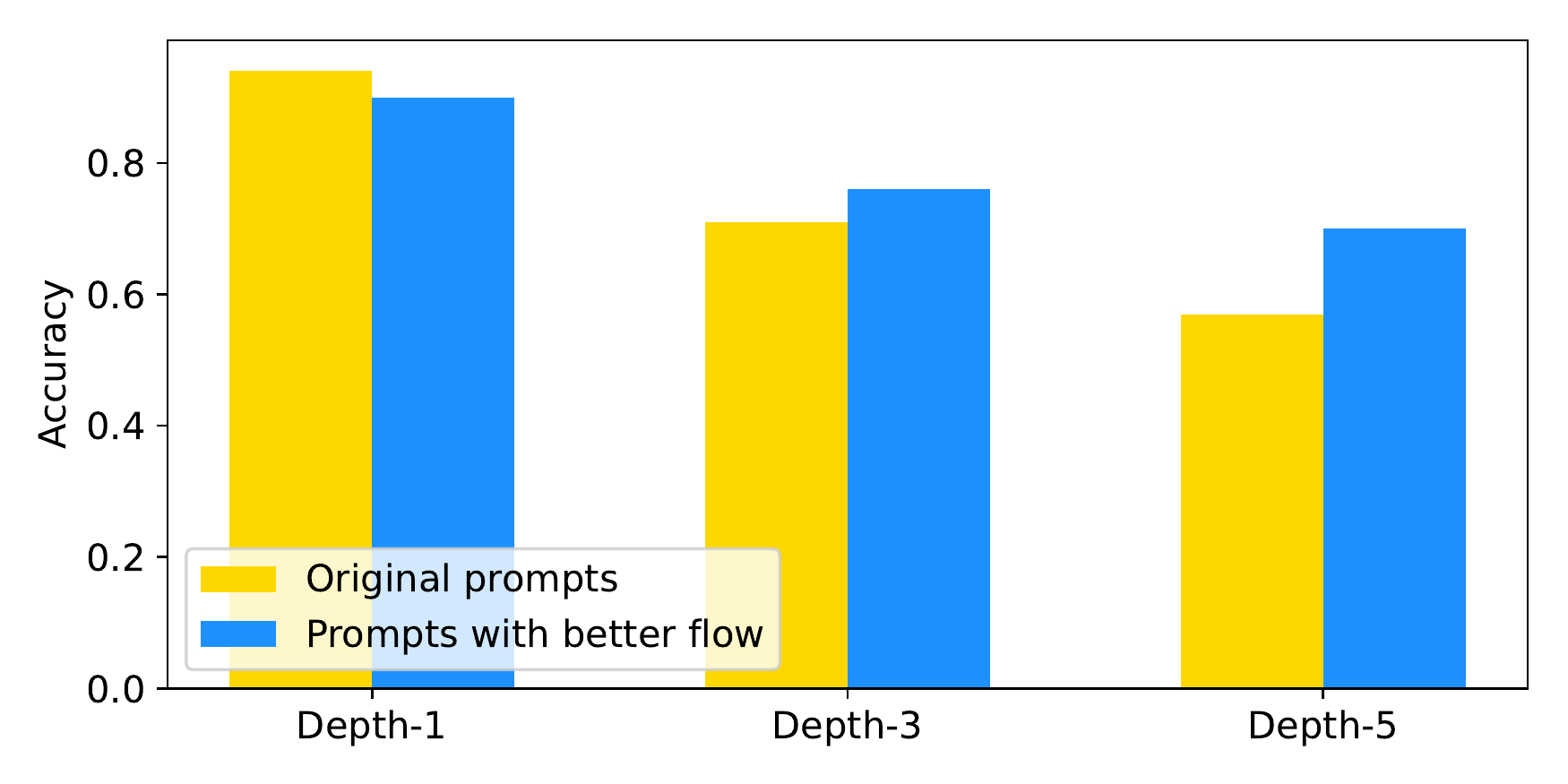}
  \caption{%
  \label{fig:cot-prompts} %
    CoT results on PrOntoQA with the original prompts vs. the prompts with conjunctive words added to make the sentences flow better.
  }
\end{figure}

For our experiments, we used the PaLM 540B model \cite{chowdhery2022palm} for all the models (both \algo\ and the baselines) served on a $4\times4$ TPU v4 architecture. The decoding temperature was set to zero. For testing CoT on PrOntoQA, we used the same demonstration examples as the original work but slightly changed the wording by adding conjunctive words such as ``\texttt{Since}'' and ``\texttt{So}'' to make the chains have a better flow. The reason for this modification was that we found when working with PaLM, prompts that have a better flow result in better predictions. This can be viewed from Figure~\ref{fig:cot-prompts} where we compare the performance for the original prompts vs. the prompts with the conjunctive words added. It can be viewed that while the latter slightly underperforms on Depth-1 (where the reasoning flow is not as important), it substantially improves the results for higher depths (especially Depth-5). For ProofWriter, we wrote similar few-shot examples. 

For SI, we used the same demonstration examples as in the original work for ProofWriter; for PrOntoQA we wrote few-shot examples following a similar pattern to those for ProofWriter. For each dataset depth we used/wrote specific few-shot examples (e.g., when working with a subset of the data that has examples requiring at most $k$ hops of reasoning, our CoT demonstrations also require only $k$ hops of reasoning), except for ProofWriter Depth-5 where, following the original work, we used it for testing length-generalization and only included examples with chains up to $3$ hops. For running CoT on ProofWriter-PUD, we included extra few-shot examples where the label is \unk; the explanation for these examples is that the goal cannot be proved or disproved with a combination of the facts and the rules. For running SI on ProofWriter-PUD, after obtaining the inferences by running SI, we give the inferences and the goal to our \module{Fact Check} module which decides if the goal can be proved, disproved, or neither. Since \emph{ProofWriter-PD} and \emph{PrOntoQA} are binary datasets but \algo\  makes three-way predictions (\proved, \disproved, and \unk), to test \algo\ on these datasets, similar to SI we combine the \unk\ and \disproved\ predictions into one class.

\subsection{Datasets for Individual Module Evaluation} \label{sec:individual-datasets}
For creating datasets for measuring the performance of individual modules in \algo, we proceeded as follows. For \module{Fact Check}, we randomly selected 100 examples from the Depth-0 examples. 
We count a model prediction to be correct if it produces the same label as the one specified in the ProofWriter dataset. 
For \module{Rule Selection}, we randomly selected 100 examples and manually enumerated every rule whose consequent unifies with the goal. A model prediction is considered correct if it predicts \emph{all} such rules correctly. For \module{Goal Decomposition}, we randomly selected 100 rules and goals such that the consequent of the rule unifies with the goal and then manually wrote the sub-goals. A model prediction is considered correct if it predicts \emph{all} the sub-goals correctly. For \module{Sign Agreement}, we re-used the same examples from the \module{Goal Decomposition} module and manually labeled them with respect to their sign agreement/disagreement.

\subsection{Quality Issues in ParaRules} \label{sec:natlang-preproc}
We found the ParaRules dataset to has a high amount of variation in the text, in the facts, and in the rules thus making it a valuable benchmark for evaluating text-based logical reasoning.
We also found a few quality issues in the ParaRules dataset that were introduced when annotators converted facts and rules into natural language form. Here, we describe some of the main issues that we found and fixed. 
\begin{itemize} [nosep,leftmargin=3.5mm]
    \item \textbf{Changing antecedents and consequents:} We found that in some cases where the rule was \texttt{``X and Y imply Z''}, the natural language version of the rule produced by annotators was written as if \texttt{``X implies Y and Z''} or \texttt{``X implies Y or Z''}. As an example, the rule \texttt{``Cold, nice people are red.''} was written in natural language form as \texttt{``Some cold people can be nice at times,and red at at other times.''}. For such cases, we modified the text to make the antecedents and consequent match the original rule.
    \item \textbf{Introducing new antecedents:} In some cases, the annotator introduced new antecedents in the rule. For example, for a rule where the antecedents were \texttt{``green''}, \texttt{``red''} and \texttt{``rough''}, the annotator added another antecedent \texttt{``naive''} (\texttt{``If someone is green and naive ...''}). For such cases, we removed the extra antecedents.
    \item \textbf{Turning general rules to specific ones:} In some cases, the natural language version of a general rule was written for only a specific entity. For example the rule \texttt{``Rough, young, green people are very round.''} was written as \texttt{``Tom is a rough, young person to know ...''}. We removed the specific entities and made the rule generally applicable.
    \item \textbf{Introducing pronouns:} For some of the facts, we found that the annotator replaced the name of the entity with a pronoun. As an example, \texttt{``Dave is ...''} was annotated as \texttt{``He is ...''}. We replaced the pronouns with the original entity name in the theory.
\end{itemize}

\begin{algorithm}[t]
\caption{FactCheck}
\label{algo:fact-check}
\textbf{Input:} Facts $\facts{F}$, Goal $\goal{G}$, Number of trials $n$
\begin{algorithmic}[1]
\FOR{$n$ times do}
    \STATE f = \textcolor{blue}{FactSelection}($\facts{F}$, $\goal{G}$)
    \STATE result = \textcolor{blue}{FactVerifier}(f, $\goal{G}$)
    \IF{result $\neq$ \unk}
        \STATE \textbf{return} result
    \ENDIF 
    \STATE $\facts{F} = \facts{F} - f$
\ENDFOR

\STATE \textbf{return} \unk
\end{algorithmic}
\end{algorithm}

\begin{algorithm}[t]
\caption{RuleSelection}
\label{algo:rule-selection}
\textbf{Input:} Rules $\rules{R}$, Goal $\goal{G}$
\begin{algorithmic}[1]
\STATE $\mathbf{I}$ = \textcolor{blue}{RuleImplications}($\rules{R}$)
\STATE selected = \textcolor{blue}{SelectRules}($\mathbf{I}$, $\goal{G}$)
\STATE \textbf{return} selected
\end{algorithmic}
\end{algorithm}

\subsection{Prompts}
\label{sec:prompts}
We provide an overview of the prompts we used for each of the four components of our model for the ProofWriter dataset.

The pseudo-code for the \module{Fact Check} module is provided in Algorithm~\ref{algo:fact-check}. For selecting a fact in \module{Fact Check}, our prompt looks like the following:

\indent \texttt{Example 1}\\
\indent \texttt{Fact1: <FACT1> Fact2: <FACT2> ... Factn: <FACTn>}\\
\indent \texttt{Question: <QUESTION>}\\
\indent \texttt{Inference: For the question <QUESTION> the most relevant fact is Facti (<FACTi>).}\\
\indent \texttt{...}\\
\indent \texttt{Example K}\\
\indent \texttt{Fact1: <FACT> Fact2: <FACT> ... Factm: <FACT>}\\
\indent \texttt{Question: <QUESTION>}\\
\indent \texttt{Inference: }

For verifying if the goal/question can be derived from the selected fact, we use the following prompt:

\indent \texttt{Example 1}\\
\indent \texttt{Fact: <FACT>}\\
\indent \texttt{Question: <QUESTION>}\\
\indent \texttt{Inference: The fact <FACT> [X1] the question <QUESTION> so [X2].}\\
\indent \texttt{...}\\
\indent \texttt{Example K}\\
\indent \texttt{Fact: <FACT>}\\
\indent \texttt{Question: <QUESTION>}\\
\indent \texttt{Inference: }\\

In the case where the goal can be proved from the fact, we replace \texttt{[X1]} with \texttt{``is equivalent to''} and \texttt{[X2]} with \texttt{``so the answer is "yes"''}. In the case where the goal can be disproved from the fact, we replace \texttt{[X1]} with \texttt{``is the negation of''} and \texttt{[X2]} with \texttt{``so the answer is "no"''}. And in the case where the goal can neither be proved nor disproved, we replace \texttt{[X1]} with \texttt{``is neither equivalent nor the negation of''} and \texttt{[X2]} with \texttt{``so the question cannot be inferred from the fact''}.

The pseudo-code for the \module{Rule Selection} module is provided in Algorithm~\ref{algo:rule-selection}.
For finding the implication/consequent of the rules, we use the following prompt:

\indent \texttt{Example 1}\\
\indent \texttt{Rule1: <RULE1>, Rule2: <RULE2> ... Rulen: <RULEn>}\\
\indent \texttt{Inference: Rule1 implies [X1], $\dots$, Rulen implies [Xn].}\\
\indent \texttt{...}\\
\indent \texttt{Example K}\\
\indent \texttt{Rule1: <RULE1>, Rule2: <RULE2> ... Rulem: <RULEm>}\\
\indent \texttt{Inference: }\\

\texttt{[Xi]}s depend on the consequent of each rule. For rules such as \texttt{``Rough, nice people are red.''} we write \texttt{[Xi]} as \texttt{``(is; red)''}, and for rules such as \texttt{``If the cat chases the dog then the cat sees the dog.''} we write \texttt{[Xi]} as \texttt{``(cat; chase; dog)''}.

For rule selection based on the implications, we use the following prompt:

\indent \texttt{Example 1}\\
\indent \texttt{Rule1 implies <IMLP1>, Rule2 implies <IMPL2>, ..., Rulen implies <IMPLn>}\\
\indent \texttt{Question: <QUESTION>}\\
\indent \texttt{Inference: The question is about <IMPLq>: Rule1 <IMPL1> [X1] <IMPLq>, $\dots$, <IMPLn> [Xn] <IMPLq>.}\\
\indent \texttt{...}\\
\indent \texttt{Example K}\\
\indent \texttt{Rule1 implies <IMLP1>, Rule2 implies <IMPL2>, ..., Rulem implies <IMPLm>}\\
\indent \texttt{Question: <QUESTION>}\\
\indent \texttt{Inference: }\\

where each \texttt{[X1]} is either \texttt{``is applicable to``} or \texttt{``not applicable to``} depending on whether the rule can be applied or not.

For goal decomposition, we use the following prompt:

\indent \texttt{Example 1}\\
\indent \texttt{Rule: <Rule>}\\
\indent \texttt{Question: <QUESTION>}\\
\indent \texttt{Inference: The question subject is <SUBJq> and the rule premises are <PRM>*, so the question breaks down to <SUBQ>*.}\\
\indent \texttt{...}\\
\indent \texttt{Example K}\\
\indent \texttt{Rule: <RULE>}\\
\indent \texttt{Question: <QUESTION>}\\
\indent \texttt{Inference: }\\

where \texttt{<SUBJq>} indicates the subject of the question, \texttt{<PRM>*} indicates the premises/antecedents in the rule (the * indicates that there might be multiple premises), and \texttt{<SUBQ>*} indicates the sub-goals.

Finally, for sign agreement, we use the following prompt:

\indent \texttt{Example 1}\\
\indent \texttt{Rule: <Rule>}\\
\indent \texttt{Question: <QUESTION>}\\
\indent \texttt{Inference: The rule implication <IMLPr> is [Xr], the question <IMPLq> is [Xq], so signs [Xd].}\\
\indent \texttt{...}\\
\indent \texttt{Example K}\\
\indent \texttt{Rule: <RULE>}\\
\indent \texttt{Question: <QUESTION>}\\
\indent \texttt{Inference: }\\

where \texttt{<IMLPr>} shows the implication of the rule and \texttt{<IMPLq>} indicates the implication of the question. \texttt{[Xr]} and \texttt{[Xq]} are either \texttt{``positive``} or \texttt{``negated``} depending on the sign of the implication. \texttt{[Xd]} is either \texttt{``agree``} or \texttt{``disagree``} depending on whether the signs agree or not.

\end{document}